\documentclass[11pt, a4paper, nonumbering]{llava}
\usepackage{enumitem}
\usepackage[authoryear, compress, round]{natbib}
\usepackage{dblfloatfix}
\usepackage{ulem}
\usepackage{caption}
\usepackage{dramatist}
\usepackage{xspace}
\usepackage{pifont} %
\usepackage{multirow}
\usepackage{tcolorbox}
\usepackage{xltabular}
\usepackage{longtable}
\usepackage{hyperref}
\usepackage{booktabs}

\usepackage{wrapfig}
\usepackage{graphicx}
\usepackage[table]{xcolor} 
\usepackage{siunitx}
\usepackage{threeparttable}
\usepackage{lmodern}
\usepackage{amsfonts}
\usepackage{amsmath}
\usepackage{amssymb}
\usepackage{lineno} 
\usepackage{multirow}
\usepackage{tikz}
\usepackage[bottom]{footmisc}
\usepackage[utf8]{inputenc}   
\usepackage{CJKutf8}
\usepackage{subfigure}
\usepackage{setspace}

\usepackage{makecell}

\usepackage{graphicx}
\usepackage{subcaption}
\usepackage{multicol} %
\usepackage{siunitx}  %
\usepackage{wrapfig}

\definecolor{HeaderBG}{HTML}{F8F9FB}    
\definecolor{RowAlt}{HTML}{FDFDFE}      
\definecolor{BorderGray}{HTML}{E1E5E9}  

\usepackage{fontawesome}
\usepackage{xspace}
\newcommand{\huggingface}{\raisebox{-1.5pt}{\includegraphics[height=1.05em]{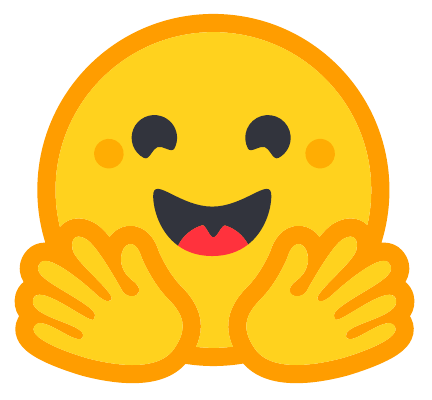}}\xspace}
\newcommand{\github}{\raisebox{-1.5pt}{\includegraphics[height=1.05em]{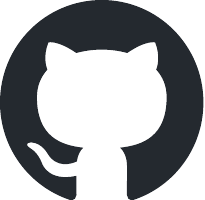}}\xspace}

\def\vlname{LLaVA-OneVision-1.5}
\def\money{\textit{\$16,000}}
\def\pretraindb{LLaVA-OneVision-1.5-Mid-Traning}
\def\instructdb{LLaVA-OneVision-1.5-Instruct}

\title{\centering \textit{LLaVA-OneVision-1.5}: Fully Open Framework for Democratized Multimodal Training}
\author{
 \textbf{LLaVA-OneVision Community Contributors}
}

\begin{abstract}
We present \vlname, a novel family of Large Multimodal Models (LMMs) that achieve state-of-the-art performance with significantly reduced computational and financial costs. Different from the existing works, \vlname\ provides an open, efficient, and reproducible framework for building high-quality vision-language models entirely from scratch. The \vlname\ release comprises three primary components: \textit{\textbf{(1) Large-Scale Curated Datasets:}} We construct an 85M concept-balanced pretraining dataset \pretraindb\ and a meticulously curated 22M instruction dataset \instructdb. \textit{\textbf{(2) Efficient Training Framework:}} We develop a complete end-to-end efficient training framework leveraging an offline parallel data packing strategy to facilitate the training of \vlname\ within a \money\ budget. \textit{\textbf{(3) State-of-the-art Performance:}} Experimental results demonstrate that \vlname\ yields exceptionally competitive performance across a broad range of downstream tasks. Specifically, \vlname-8B outperforms Qwen2.5-VL-7B on 18 of 27 benchmarks, and \vlname-4B surpasses Qwen2.5-VL-3B on all 27 benchmarks. (4) \textit{\textbf{RL-based Post-training:}} We unlock the model's latent potential through a lightweight RL stage, effectively eliciting robust chain-of-thought reasoning to significantly boost performance on complex multimodal reasoning tasks. 

\end{abstract}

\begin{document}
\maketitle
\vspace{-0.5cm}
\begin{center}
    \renewcommand{\arraystretch}{1}
    \resizebox{\linewidth}{!}{
    \begin{tabular}{lll}
        \github & \textbf{Code} & {\url{https://github.com/EvolvingLMMs-Lab/LLaVA-OneVision-1.5}} \\
        \github & \textbf{RL Code} & {\url{https://github.com/EvolvingLMMs-Lab/LLaVA-OneVision-1.5-RL}} \\
        \huggingface & \textbf{Models} & \url{https://huggingface.co/lmms-lab/LLaVA-OneVision-1.5-8B-Instruct}\\
        \huggingface & \textbf{RL Model} & \url{https://huggingface.co/mvp-lab/LLaVA-OV-1.5-8B-RL}\\
        \huggingface & \textbf{Pretrain data} & \url{https://huggingface.co/datasets/mvp-lab/LLaVA-OneVision-1.5-Mid-Training-85M}\\
        \huggingface & \textbf{Instruct data} & \url{https://huggingface.co/datasets/mvp-lab/LLaVA-OneVision-1.5-Instruct-Data}\\
        \huggingface & \textbf{RL Data} & \url{https://huggingface.co/datasets/mvp-lab/LLaVA-OneVision-1.5-RL-Data}\\
    \end{tabular}
    }
\end{center}


\begin{figure}[htbp]
    \centering
     \includegraphics[width=0.52\textwidth]{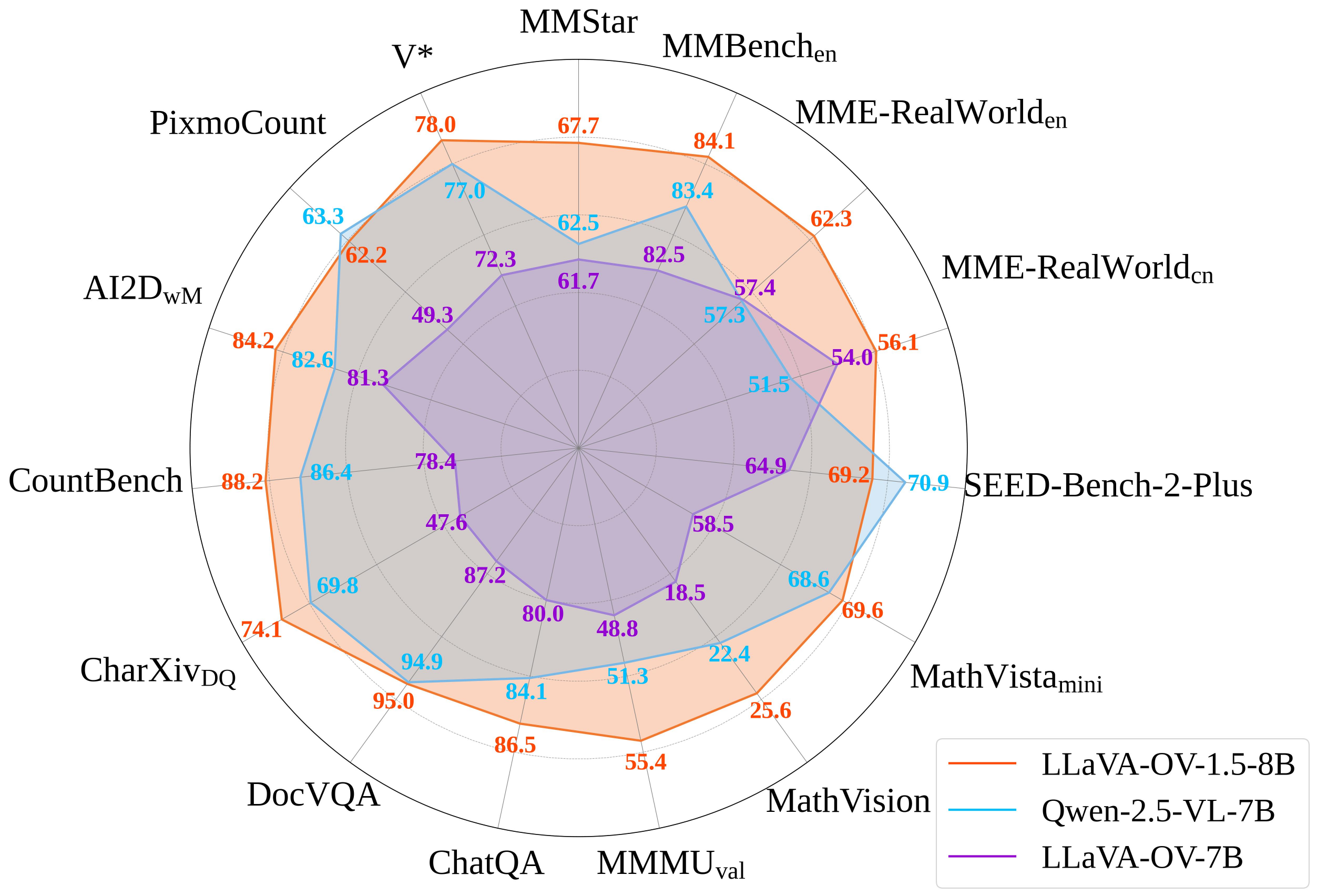}
    \hfill
    \includegraphics[width=0.46\textwidth]{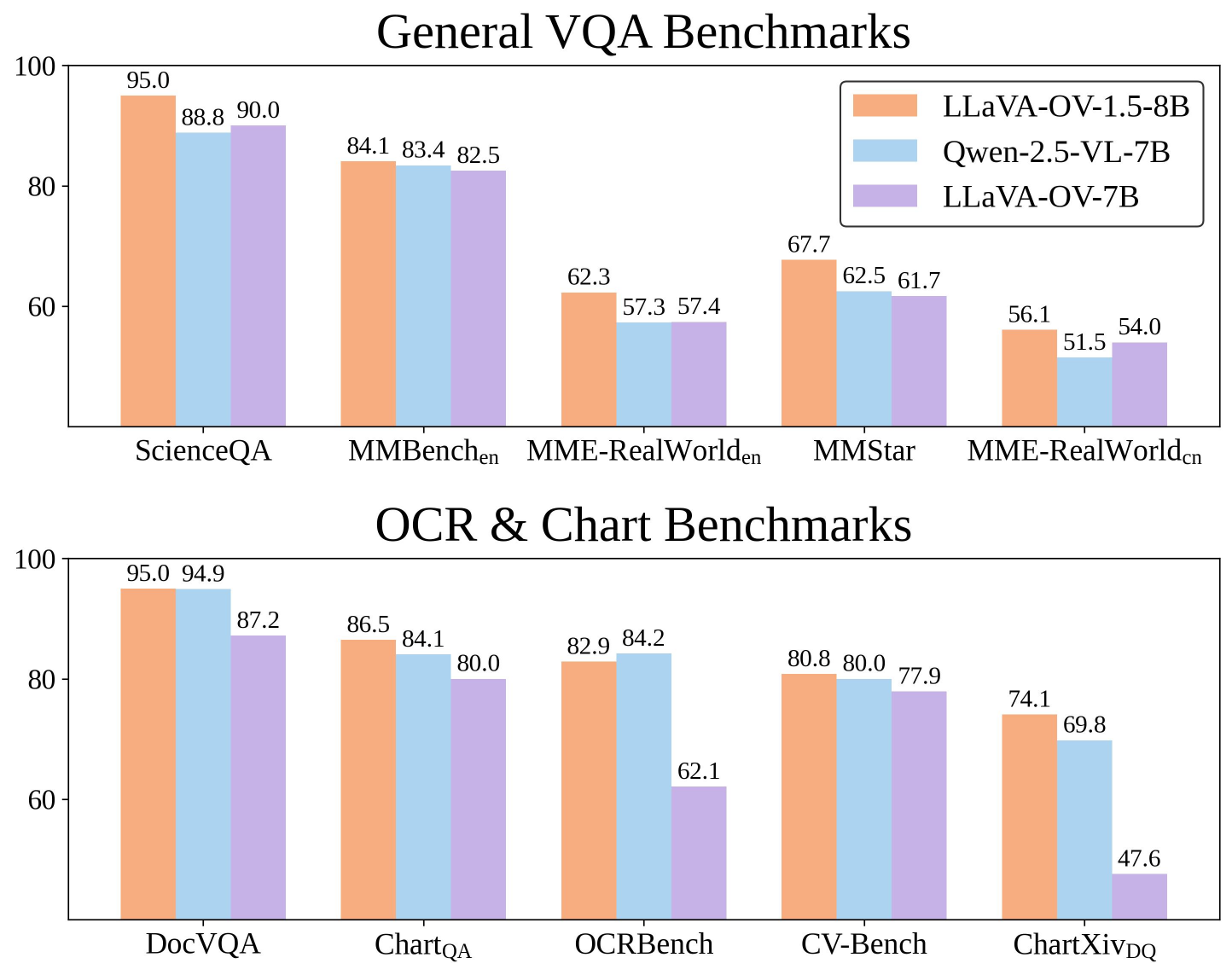}
    \caption{
    Performance of LLava-OV-1.5-8B across multiple benchmarks.
    }
    \label{fig:mimo_comparison}
\end{figure}

\section{Introduction}
Recent advancements in Large Multimodal Models~(LMMs) have demonstrated remarkable capabilities in multimodal understanding and reasoning~\citep{gemini2.5,seedv11.5,internvl3}. These developments enable artificial intelligence applications to effectively comprehend and analyze images, charts, and PDF documents. 
However, the most performant models remain proprietary, with neither their training data nor source code publicly available. Consequently, the broader research community lacks crucial insights into how such high-performing LMMs can be built from scratch.

To reduce barriers for community development, several research efforts have attempted to reproduce the capabilities of proprietary models using open architectures. Early efforts such as LLaVA~\citep{llava}, LLaVA-Next~\citep{llavanext}, and LLaVA-OneVision~\citep{llavaov} provided fully open training data and code, but their performance now falls substantially behind that of current state-of-the-art models~\citep{qwen2.5vl,internvl3}. More recent works have pushed the boundary: Molmo~\citep{molmo} released model weights, datasets, and source code, enabling the community to train LMMs from scratch. Through careful architectural choices, a refined training pipeline, and high-quality data, Molmo achieves near-parity with GPT-4V on both academic benchmarks and user preference evaluations. Open-Qwen2VL~\citep{wang2025open} introduces a 2B-parameter model pre-trained on only 0.36\% of 1.4T multimodal tokens in Qwen2-VL, while outperforming Qwen2-VL-2B across various multimodal benchmarks. Despite these advances, the performance gap between open-source and proprietary models continues to widen as the field of LMMs rapidly evolves. Current open-source models are still constrained by substantial computational demands and suboptimal training efficiency.

To overcome the aforementioned limitations, we introduce \vlname, a fully open-source family of LMMs, extending the LLaVA series~\citep{llavaov} to achieve superior performance with limited computational cost. Specifically, \vlname\ adopts RICE-ViT~\citep{rice} as the vision encoder, enabling native-resolution adaptation and fine-grained visual understanding based on stronger region-level semantic representation. Building upon LLaVA-OneVision~\citep{llavaov}, \vlname\ adopts a three-stage training pipeline: (Stage-1) Language-Image Alignment, (Stage-1.5) High-Quality Knowledge Learning, and (Stage-2) Visual Instruction Tuning. Notably, we find that simply scaling data at the mid-training stage alone can produce state-of-the-art LMMs, eliminating the need for complex training paradigms. To foster open research, we release all assets to the community, including \pretraindb\ and \instructdb\ datasets, the training framework, and model checkpoints (\vlname-Base and \vlname-Instruct). In summary, our contributions are as follows:

\begin{itemize}[leftmargin=*]
    \item \textit{\textbf{Large Multimodal Models.}} We propose \vlname\, a family of fully open-source large multimodal models that achieve superior performance across multiple multimodal benchmarks compared to Qwen2.5-VL.
    \item \textit{\textbf{Large-Scale Datasets.}} We construct an 85M concept-balanced pre-training dataset \pretraindb\ and a meticulously curated 22M instruction dataset \instructdb.
    \item \textit{\textbf{Efficienct Training Framework.}} We develop a complete end-to-end training framework that employs an offline parallel data packing strategy to optimize cost-effectiveness, enabling the training of \vlname\ within a \money\ compute budget.
    \item \textit{\textbf{Open-Source Release.}} We release all assets to the public including \pretraindb\ and \instructdb\ dataset, training framework, and the model checkpoints (\vlname-Base and \vlname-Instruct).
    \item \textit{\textbf{RL-Enhanced Reasoning.}} We introduce a lightweight RL post-training stage using the asynchronous AReaL system. By adopting a discrepancy-driven data selection strategy and rigorous outcome-based verification, we effectively elicit latent reasoning capabilities, significantly boosting performance on complex tasks while maintaining robust general visual understanding.
 \end{itemize}

\section{Architecture}

\begin{figure}[t!]
    \centering
     \includegraphics[width=\textwidth]{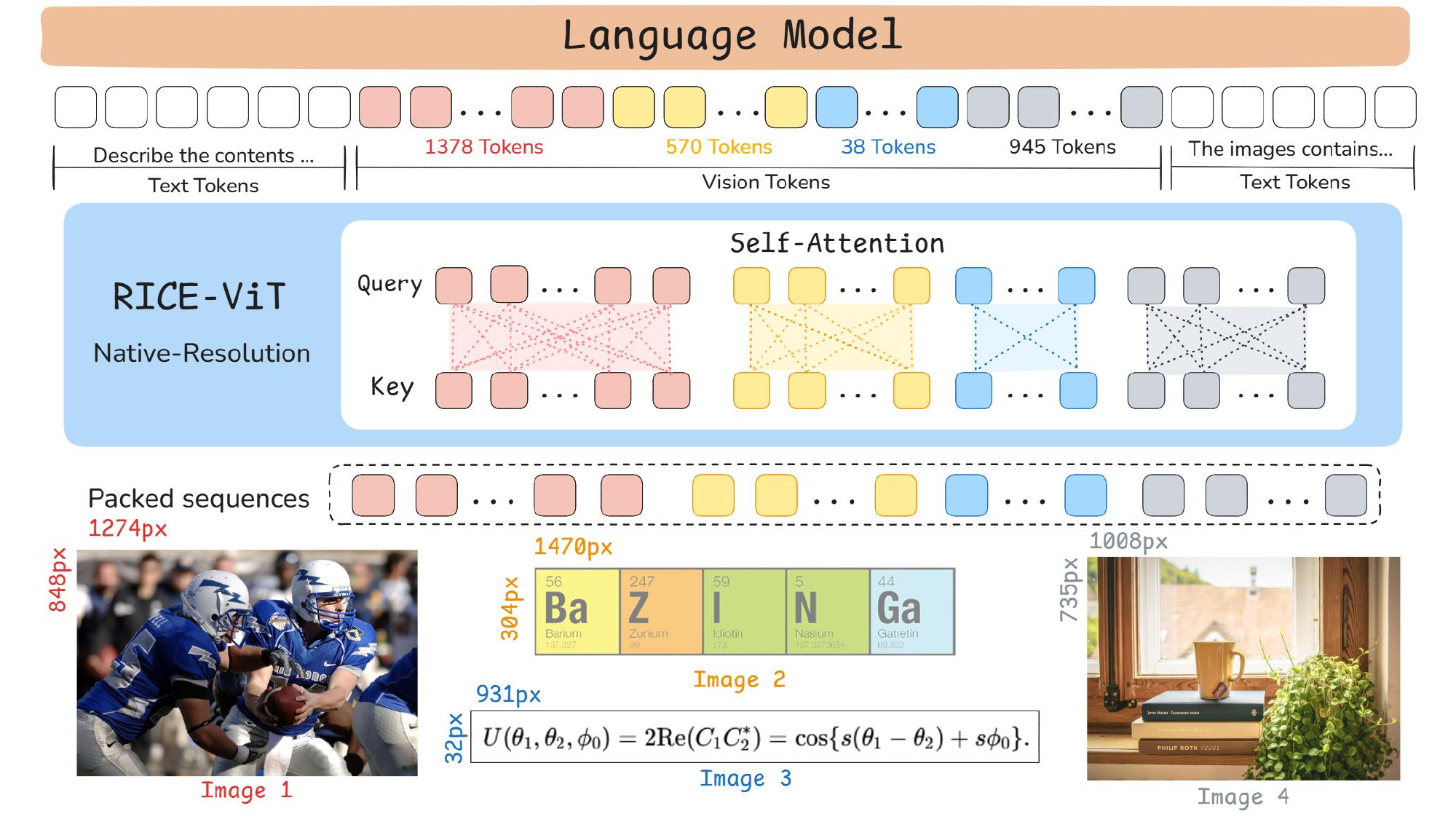}
    \vspace{-5mm}
    \caption{Overall architecture of \vlname. The framework integrates a pre-trained vision encoder with a language model decoder. The vision encoder adopts 2D RoPE for native-resolution processing and incorporates region-aware attention to enhance local semantic modeling. During pretraining, both object regions and OCR regions are jointly modeled to inject fine-grained text understanding capability. A lightweight projector maps visual features into the LLM embedding space, and the \texttt{[CLS]} token is preserved to retain global semantic capacity during multimodal alignment.
    }
    \label{fig:architecture}
\end{figure}

\subsection{Overall Architecture} 
The overall architecture of \vlname\ is illustrated in Fig.~\ref{fig:architecture}. \vlname\ retains the ``ViT–MLP–LLM'' paradigm of the LLaVA series, comprising three core modules:

\begin{itemize}
\item \textbf{\textit{Vision Encoder}}: The vision encoder is responsible for extracting rich and semantically meaningful visual representations from input images, which serve as the foundation for multimodal alignment and downstream reasoning. Unlike previous works~\citep{wang2024qwen2,qwen2.5vl} that adopt SigLIP~\citep{siglip} or DFN~\citep{dfn}, \vlname\ integrates our recently proposed cluster discrimination model RICE-ViT~\citep{rice} to improve region-aware visual and OCR capabilities.

\item  \textbf{\textit{Projector}}: The projector bridges the modality gap between the vision encoder and the large language model by mapping visual embeddings into the text embedding space of the LLM. Following Qwen2.5-VL~\citep{qwen2.5vl}, we first group spatially adjacent sets of four patch features, which are then concatenated and passed through a two-layer multi-layer perceptron to map them into the text embedding space of the LLM.

\item  \textbf{\textit{Large Language Model}}: The large language model acts as the reasoning and generation core of the architecture. After receiving the projected multimodal embeddings, the LLM integrates visual information with linguistic context to perform complex reasoning, instruction following, and natural language generation. The \vlname\ series utilize Qwen3~\citep{qwen3} as the language backbone to significantly enhance performance on downstream tasks.
\end{itemize}
This modular design follows the LLaVA framework but incorporates more efficient training recipes and carefully selected encoders, enabling superior cost-effectiveness and scalability.

\subsection{Vision Encoder via Region-Aware Cluster Discrimination}

Fine-grained visual semantics are essential for dense prediction tasks such as grounding, OCR, and segmentation. Although large-scale vision–language contrastive models like CLIP~\citep{clip} and SigLIP~\citep{siglip} demonstrate strong performance through global vision-language alignment, they fail to capture the similarity structure of training data or the local region-level semantics within images. This shortcoming stems from instance-wise contrastive learning, which treats all instances as negatives regardless of their semantic similarity and represents each instance solely with a single global embedding.

To overcome these limitations, our \vlname\ leverages the RICE-ViT~\citep{rice} as its vision encoder, enabling precise multimodal alignment and enriched region-level representation. RICE-ViT enhances both object-centric and OCR capabilities by introducing a unified region cluster discrimination loss, trained on 450M images and 2.4B candidate regions. Its design combines a region-aware attention mechanism for local semantic modeling with 2D rotary positional encoding, which naturally supports variable input resolutions without requiring resolution-specific fine-tuning, unlike models such as Qwen2-VL~\citep{wang2024qwen2} and InternVL 2.5~\citep{chen2024expanding}.

We integrate this pretrained encoder with a language model through joint training, yielding a streamlined multimodal pipeline. Compared to SigLIP2~\citep{siglipv2}, which depends on multiple specialized losses (SILC, TIPS, LocCa, and Sigmoid), our method adopts a single cluster discrimination loss that simultaneously strengthens general understanding, OCR, and localization. This unified formulation provides an elegant, computationally efficient solution that matches SigLIP2’s performance while substantially reducing architectural complexity and training overhead.
\section{Data}

\begin{figure}[t]
  \centering
  \subfigure[]{\includegraphics[width=0.36\textwidth]{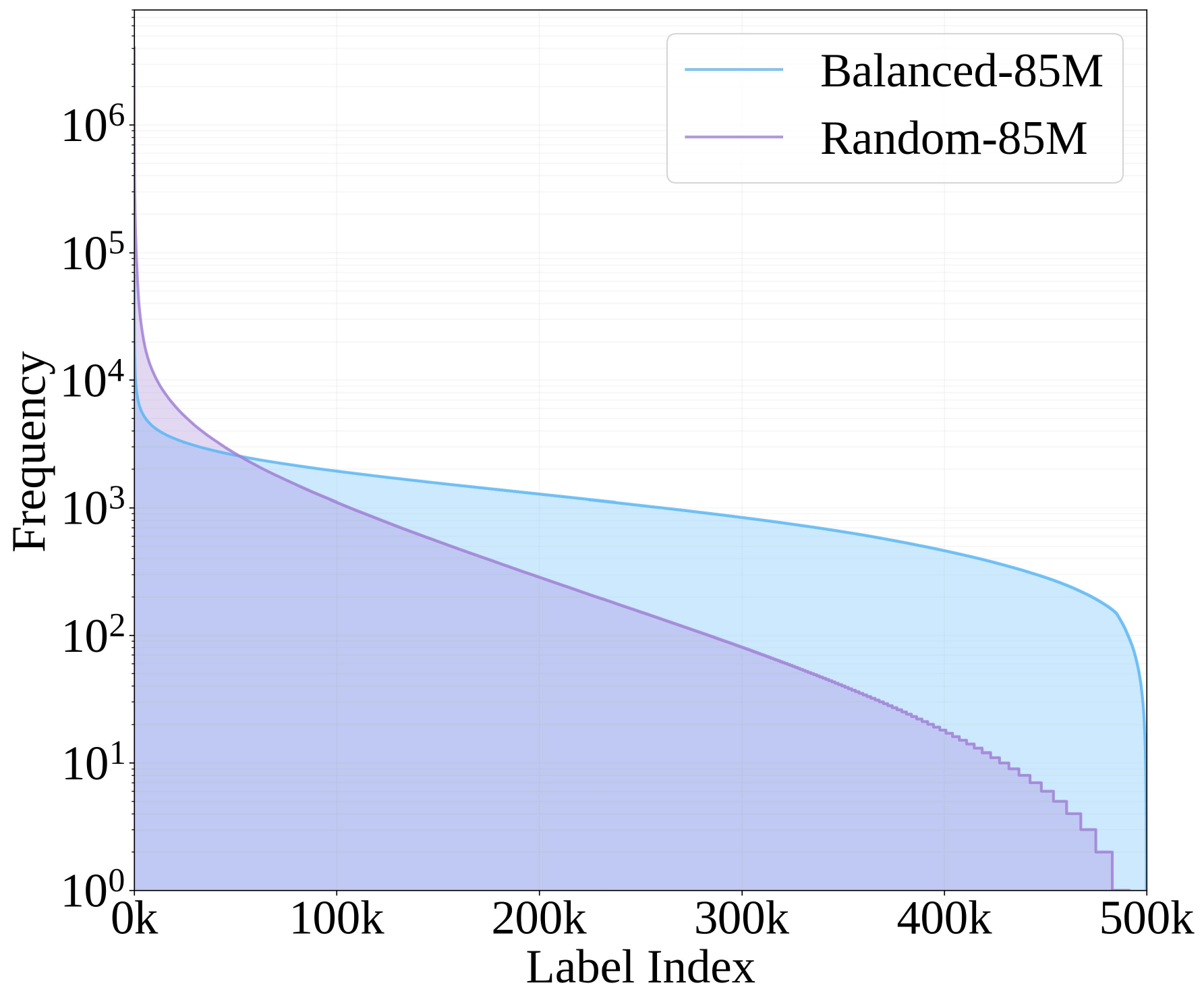}\label{fig:5a}}
  \hfill
  \subfigure[]{\includegraphics[width=0.31\textwidth]{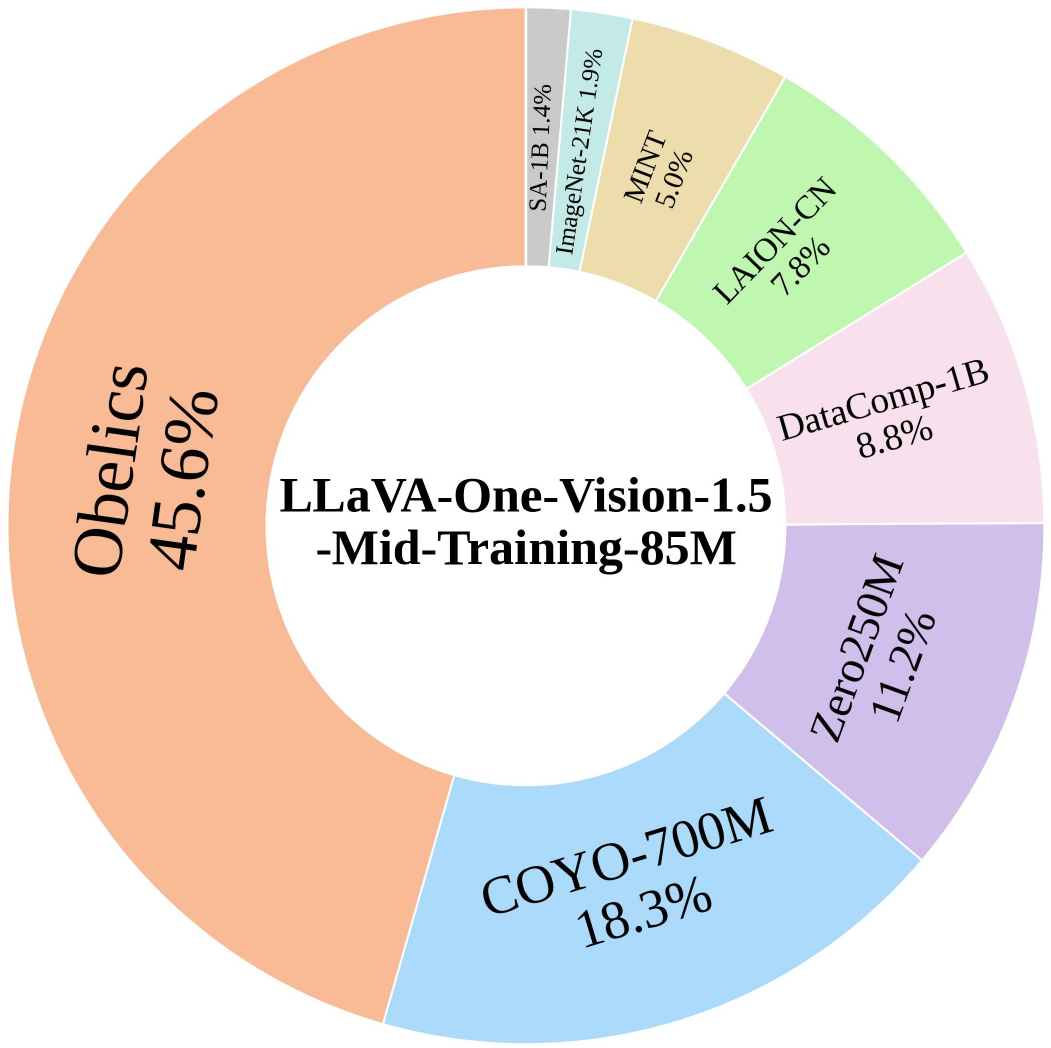}\label{fig:5b}}
  \hfill
  \subfigure[]{\includegraphics[width=0.31\textwidth]{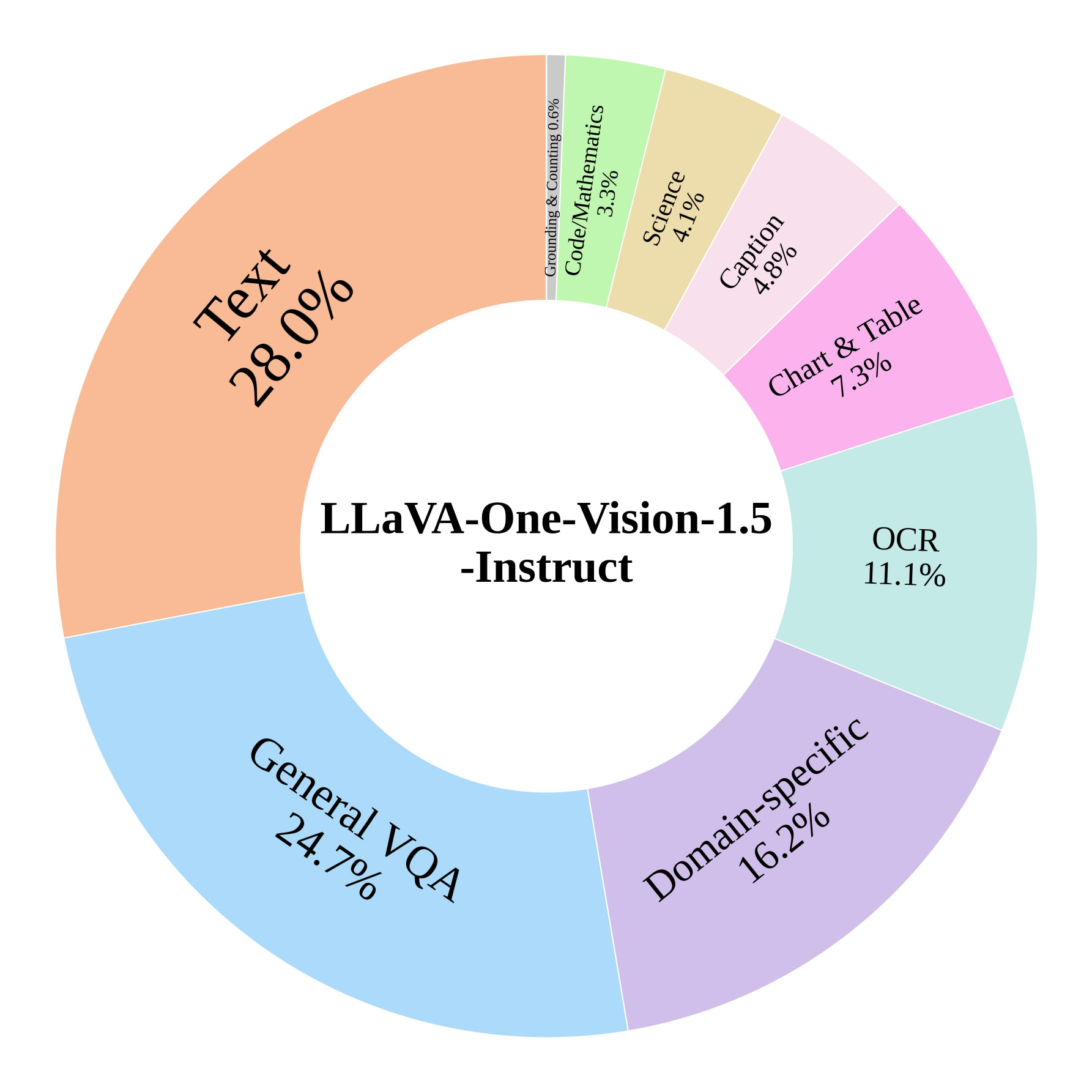}\label{fig:5c}}
  \vspace{-3mm}
  \caption{(a) The vocabulary coverage proportion in the \pretraindb\ dataset before and after concept balancing. (b) Distribution of data sources within the \pretraindb\ dataset. (c) Distribution of data sources within the \instructdb.}
  \label{fig:main}
\end{figure}

\subsection{Pre-Training Dataset}

We use the LLaVA-1.5 558K~\citep{llava1.5} to align the visual features into the word embedding space of LLMs. After that, \vlname\ is underpinned by a large-scale multimodal dataset \pretraindb, which contains 85 million high-quality image-text pairs~(20M in Chinese and 65M in English). The data of \pretraindb\ are from a wide range of sources: COYO-700M~\citep{coyo700m}, Obelics~\citep{obelics}, DataComp-1B~\citep{datacomp}, LAION-CN~\citep{laioncn}, ImageNet-21K~\citep{imagenet}, SAM-1B~\citep{sam}, MINT~\citep{mint}, and Zero250M~\citep{zero}. 
To enrich the diversity of our pretraining data, we introduce a concept-balanced sampling strategy inspired by MetaCLIP~\citep{metaclip}. Unlike MetaCLIP, which depends on raw captions for concept matching and struggles with caption-free or interleaved datasets (\textit{e.g.}, SAM-1B, ImageNet-21K, and Obelics), our method reduces reliance on caption quality, such as the brief and incomplete annotations common in COYO-700M. Instead, we adopt a feature-based matching approach that coarsely groups image sources. Specifically, using the pretrained MetaCLIP-H/14-Full-CC2.5B encoders~\citep{metaclip}, we project both images and MetaCLIP’s 500K concept entries into a shared embedding space. Since MetaCLIP embeddings are already concept-balanced, this enables effective similarity-based concept induction: for each image, we retrieve its top-K nearest concept embeddings to construct refined pseudo-captions that enhance semantic alignment.
\paragraph{Top-K Concept Assignment and Balance Sampling.}
Given an image set $\mathcal{I} = \{i_0, i_1,...,i_N\}$ and a concept vocabulary set $\mathcal{V} = \{v_0, v_1,...v_M\}$, we first utilize the image encoder $\Phi_v$ and text encoder $\Phi_t$ to extract the image embeddings $\mathcal{E}_i=\{\Phi_v \left(i\right), i \in I \}$  and concept embeddings $\mathcal{E}_t=\{\Phi_t \left(v\right), v \in V \}$. Then we assign each image with the top-$k$ nearest concepts based on the cosine similarity of the L2-normalized image and concept embeddings. After that, following MetaCLIP~\citep{metaclip}, we weight each image by the inverse frequencies of its concepts and then sample images based on the normalized image weights. This inverse-frequency methodology promotes a more balanced concept distribution, without relying on original captions, which may be noisy or missing.
This process yields 85M images with balanced concepts. Subsequently, we apply a powerful captioner to produce English and Chinese captions for these images, followed by a validity filter to eliminate duplicates and excessively lengthy outputs. Ultimately, we establish an 85M concept-balanced mid-training dataset. The distribution of data sources of the \pretraindb\ dataset is illustrated in Fig.~\ref{fig:5b}.

\subsection{Instruction Dataset}

Visual instruction tuning~\citep{llava} is vital for enabling LMMs to understand and follow visual instructions, and its effectiveness hinges on the quality of the instruction datasets. To this end, we construct the \instructdb\ dataset by aggregating a wide range of instruction-tuning datasets from diverse sources. The data are carefully curated to ensure balanced coverage across seven categories: Caption, Chart \& Table, Code \& Math, Domain-specific, General VQA, Grounding \& Counting, OCR, and Science. The resulting corpus comprises 22 million samples, with Fig.~\ref{fig:5c} showing the proportional distribution across categories.


\begin{figure}[t]
  \centering
  \subfigure[]{
    \begin{minipage}[c]{0.31\textwidth}
      \centering
      \includegraphics[width=\linewidth]{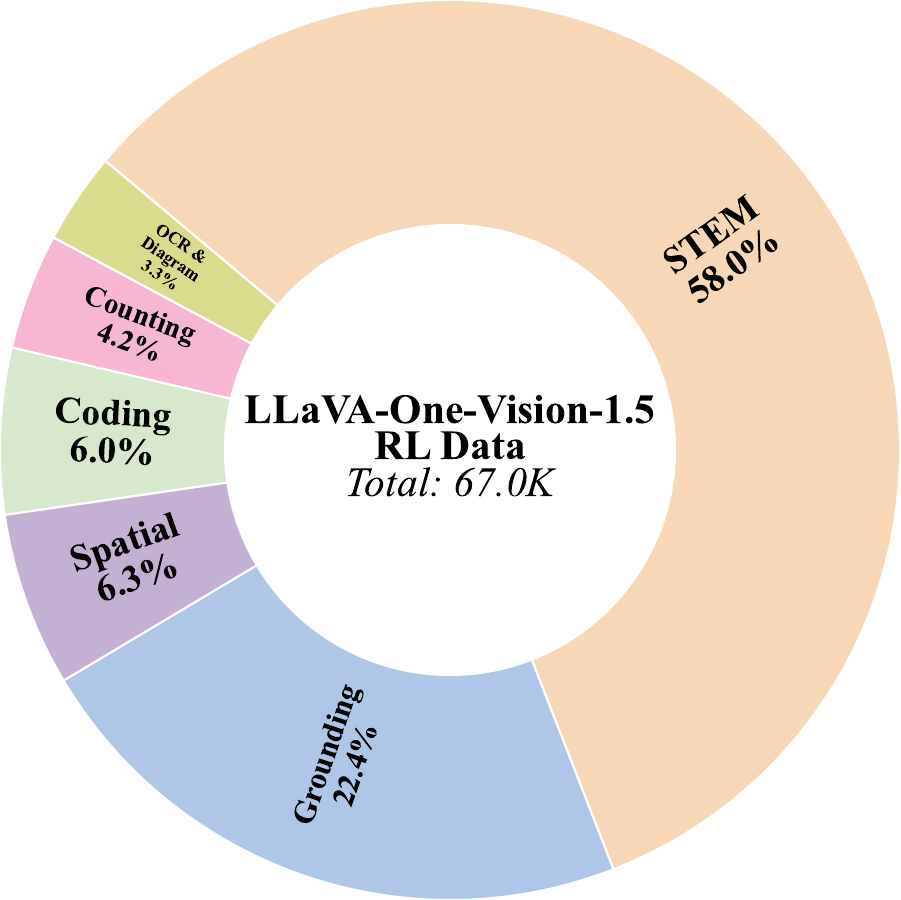}
      \label{fig:rl_stage1_pie}
    \end{minipage}
  }
  \hfill
  \subfigure[]{
    \begin{minipage}[c]{0.31\textwidth}
      \centering
      \includegraphics[width=\linewidth]{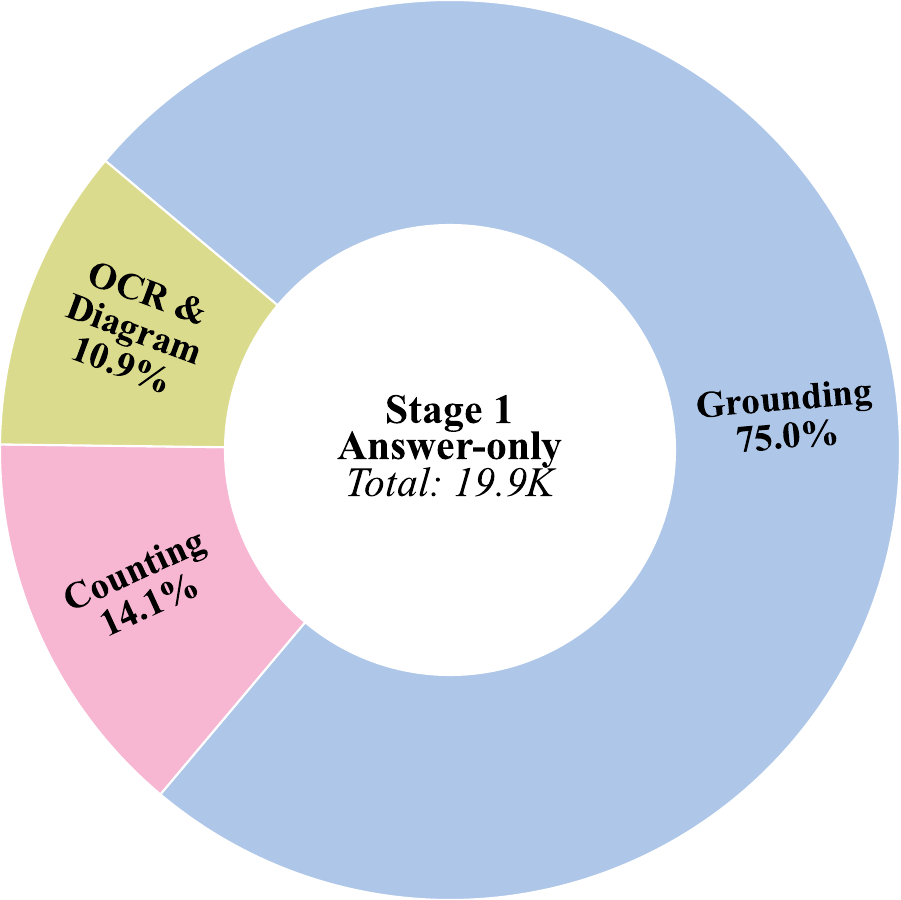}
      \label{fig:rl_stage1_pie}
    \end{minipage}
  }
  \hfill
  \subfigure[]{
    \begin{minipage}[c]{0.31\textwidth}
      \centering
      \includegraphics[width=\linewidth]{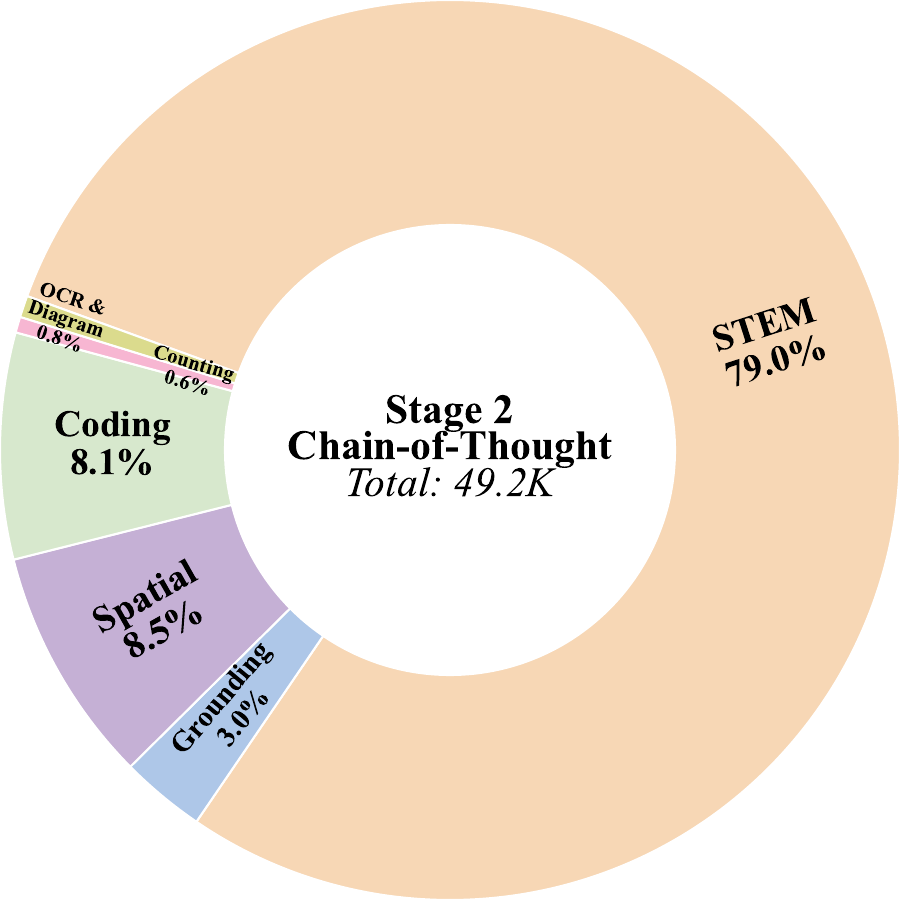}
      \label{fig:rl_stage2_pie}
    \end{minipage}
  }
  \caption{Distribution of task categories in the RL training data. (a) Total RL corpus (67K instances). (b) Stage 1: Answer-only training. (c) Stage 2: Chain-of-thought training.}
  \label{fig:rl_stage_distribution}
\end{figure}
\section{Training Strategy}

\subsection{Training Pipeline}
Following LLaVA-OneVision~\citep{llavaov}, \vlname\ undergoes three distinct learning stages to enable LLM for multimodal capabilities:
\begin{itemize}
\item \textbf{\textit{Stage-1: Language-Image Alignment.}} The stage aims to pretrain the projection layer with the LLaVA-1.5 558K to align the visual features into the word embedding space of LLMs.

\item \textbf{\textit{Stage-1.5: High-Quality Knowledge Learning.}} 
Building on the language-image alignment stage, we introduce the high-quality knowledge learning stage to strike a balance between computational efficiency and injecting new knowledge into LMMs. In this stage, we transition to full-parameter training of all modules using the \pretraindb\ dataset.

\item \textbf{\textit{Stage-2: Visual Instruction Tuning.}} To enable LMMs to handle a diverse range of visual tasks with desired responses, we continue full-parameter training with the proposed \instructdb\ as well as the FineVision~\citep{finevision2025} dataset. 

\end{itemize}

\begin{table*}[t!]
    \centering
    \definecolor{myorange}{RGB}{204,113,37}
    \resizebox{0.9\linewidth}{!}
    {
    \setlength{\tabcolsep}{6pt}
    \renewcommand{\arraystretch}{1.0}
    \small
    \begin{threeparttable}
    \caption{Performance comparison across vision-language models on various benchmarks grouped by task type. All scores are reported as accuracy percentages unless otherwise specified.}
    \label{tab:main}
    \begin{tabular}{ll
      >{\columncolor{orange!10}}S[table-format=2.1, round-mode=places, round-precision=1, round-pad=true]
      >{\columncolor{orange!10}}S[table-format=2.1, round-mode=places, round-precision=1, round-pad=true]
      >{\columncolor{orange!10}}S[table-format=2.1, round-mode=places, round-precision=1, round-pad=true]
      S[table-format=2.1, round-mode=places, round-precision=1, round-pad=true]
      >{\columncolor{orange!10}}S[table-format=2.1, round-mode=places, round-precision=1, round-pad=true]
      S[table-format=2.1, round-mode=places, round-precision=1, round-pad=true]
      S[table-format=2.1, round-mode=places, round-precision=1, round-pad=true]
      S[table-format=2.1, round-mode=places, round-precision=1, round-pad=true]}
    \toprule
    \textbf{Task} & \textbf{Benchmark} & 
    \multicolumn{1}{>{\columncolor{orange!10}}c}{\textbf{LLaVA-OV-1.5}} & 
    \multicolumn{2}{>{\columncolor{orange!10}}c}{\textbf{LLaVA-OV-1.5 RL}} & 
    \multicolumn{1}{c}{\textbf{Qwen2.5-VL}} & 
    \multicolumn{1}{>{\columncolor{orange!10}}c}{\textbf{LLaVA-OV-1.5}} & 
    \multicolumn{1}{c}{\textbf{Qwen2.5-VL}} & 
    \multicolumn{1}{c}{\textbf{LLaVA-OV}} \\
    \textbf{Size} & & 
    \multicolumn{1}{>{\columncolor{orange!10}}c}{\textbf{8B}} & 
    \multicolumn{2}{>{\columncolor{orange!10}}c}{\textbf{8B}} & 
    \multicolumn{1}{c}{\textbf{7B}} & 
    \multicolumn{1}{>{\columncolor{orange!10}}c}{\textbf{4B}} & 
    \multicolumn{1}{c}{\textbf{3B}} & 
    \multicolumn{1}{c}{\textbf{7B}} \\
    \textbf{Mode} & & 
    \multicolumn{1}{>{\columncolor{orange!10}}c}{-} & 
    \multicolumn{1}{>{\columncolor{orange!10}}c}{\textbf{thinking}} & 
    \multicolumn{1}{>{\columncolor{orange!10}}c}{\textbf{fast}} & 
    \multicolumn{1}{c}{-} & 
    \multicolumn{1}{>{\columncolor{orange!10}}c}{-} & 
    \multicolumn{1}{c}{-} & 
    \multicolumn{1}{c}{-} \\
    \midrule
    \multirow{11}{*}{\makecell[l]{General VQA}} 
     &MMStar                     & ${67.7}^{\phantom{{\uparrow0.0}}}$ & ${68.2}^{{\textcolor{myorange}{\uparrow0.5}}}$ & $\textbf{68.3}^{{\textcolor{myorange}{\uparrow0.6}}}$ & ${62.5}^{\phantom{\textcolor{myorange}{\uparrow0.0}}}$ & ${64.9}^{\phantom{\textcolor{myorange}{\uparrow0.0}}}$ & ${55.9}^{\phantom{\textcolor{myorange}{\uparrow0.0}}}$ & ${61.7}^{\phantom{\textcolor{myorange}{\uparrow0.0}}}$ \\
     &MMBench$_{\text{en}}$ & ${84.1}^{\phantom{{\uparrow0.0}}}$ & $\textbf{85.7}^{\textcolor{myorange}{\uparrow1.6}}$ & ${85.7}^{\textcolor{myorange}{\uparrow1.6}}$ & ${83.4}^{\phantom{\textcolor{myorange}{\uparrow0.0}}}$ & ${84.2}^{\phantom{\textcolor{myorange}{\uparrow0.0}}}$ & ${78.0}^{\phantom{\textcolor{myorange}{\uparrow0.0}}}$ & ${82.5}^{\phantom{\textcolor{myorange}{\uparrow0.0}}}$ \\
     &MMBench$_{\text{cn}}$ & ${81.0}^{\phantom{{\uparrow0.0}}}$ & $\textbf{84.2}^{\textcolor{myorange}{\uparrow3.2}}$ & ${81.5}^{\textcolor{myorange}{\uparrow0.5}}$ & ${81.6}^{\phantom{\textcolor{myorange}{\uparrow0.0}}}$ & ${76.9}^{\phantom{\textcolor{myorange}{\uparrow0.0}}}$ & ${74.6}^{\phantom{\textcolor{myorange}{\uparrow0.0}}}$ & ${81.4}^{\phantom{\textcolor{myorange}{\uparrow0.0}}}$ \\
     &MME-RealWorld$_{\text{en}}$ & ${61.7}^{\phantom{\uparrow0.0}}$ & $\textbf{63.4}^{\textcolor{myorange}{{\uparrow1.7}}}$ & ${63.3}^{{\textcolor{myorange}{\uparrow1.6}}}$ & ${57.3}^{\phantom{\textcolor{myorange}{\uparrow0.0}}}$ & ${61.6}^{\phantom{\textcolor{myorange}{\uparrow0.0}}}$ & ${51.6}^{\phantom{\textcolor{myorange}{\uparrow0.0}}}$ & ${57.4}^{\phantom{\textcolor{myorange}{\uparrow0.0}}}$ \\
     &MME-RealWorld$_{\text{cn}}$ & ${56.1}^{\phantom{{\uparrow0.0}}}$ & ${56.1}^{\textcolor{myorange}{\uparrow0.0}}$ & $\textbf{56.3}^{\textcolor{myorange}{\uparrow0.2}}$ & ${51.5}^{\phantom{\textcolor{myorange}{\uparrow0.0}}}$ & ${49.6}^{\phantom{\textcolor{myorange}{\uparrow0.0}}}$ & ${45.4}^{\phantom{\textcolor{myorange}{\uparrow0.0}}}$ & ${54.0}^{\phantom{\textcolor{myorange}{\uparrow0.0}}}$ \\ 
     &SeedBench$_{\text{image}}$ & ${77.3}^{\phantom{{\uparrow0.0}}}$ & ${76.7}^{\phantom{{\uparrow0.0}}}$ & $\textbf{77.6}^{\textcolor{myorange}{\uparrow0.3}}$ & $\textbf{77.5}^{\phantom{\textcolor{myorange}{\uparrow0.0}}}$ & ${76.6}^{\phantom{\textcolor{myorange}{\uparrow0.0}}}$ & ${74.8}^{\phantom{\textcolor{myorange}{\uparrow0.0}}}$ & ${75.4}^{\phantom{\textcolor{myorange}{\uparrow0.0}}}$ \\
     &CV-Bench                    & ${80.7}^{\phantom{{\uparrow0.0}}}$ & $\textbf{82.9}^{\textcolor{myorange}{\uparrow2.2}}$ & ${81.1}^{{\textcolor{myorange}{\uparrow0.4}}}$ & ${80.0}^{\phantom{\textcolor{myorange}{\uparrow0.0}}}$ & ${77.2}^{\phantom{\textcolor{myorange}{\uparrow0.0}}}$ & ${71.5}^{\phantom{\textcolor{myorange}{\uparrow0.0}}}$ & ${77.9}^{\phantom{\textcolor{myorange}{\uparrow0.0}}}$ \\
     &SEED-Bench-2-Plus          & ${69.2}^{\phantom{{\uparrow0.0}}}$ & ${69.5}^{\textcolor{myorange}{\uparrow0.3}}$ & ${69.2}^{\textcolor{myorange}{\uparrow0.0}}$ & $\textbf{70.9}^{\phantom{\textcolor{myorange}{\uparrow0.0}}}$ & ${68.9}^{\phantom{\textcolor{myorange}{\uparrow0.0}}}$ & ${68.6}^{\phantom{\textcolor{myorange}{\uparrow0.0}}}$ & ${64.9}^{\phantom{\textcolor{myorange}{\uparrow0.0}}}$ \\
     &RealWorldQA                & ${68.1}^{\phantom{{\uparrow0.0}}}$ & ${68.4}^{\textcolor{myorange}{\uparrow0.3}}$ & $\textbf{70.6}^{\textcolor{myorange}{\uparrow2.5}}$ & ${68.5}^{\phantom{\textcolor{myorange}{\uparrow0.0}}}$ & ${67.8}^{\phantom{\textcolor{myorange}{\uparrow0.0}}}$ & ${60.0}^{\phantom{\textcolor{myorange}{\uparrow0.0}}}$ & ${66.3}^{\phantom{\textcolor{myorange}{\uparrow0.0}}}$ \\
     \cmidrule{2-9} \rowcolor{gray!20} & Avg. & ${71.8}^{\phantom{\textcolor{myorange}{\uparrow0.0}}}$ & $\textbf{72.8}^{\textcolor{myorange}{\uparrow1.0}}$ & ${72.6}^{\textcolor{myorange}{\uparrow0.8}}$ & ${72.2}^{\phantom{\textcolor{myorange}{\uparrow0.0}}}$ & ${72.1}^{\phantom{\textcolor{myorange}{\uparrow0.0}}}$ & ${66.4}^{\phantom{\textcolor{myorange}{\uparrow0.0}}}$ & ${71.1}^{\phantom{\textcolor{myorange}{\uparrow0.0}}}$ \\
    \midrule 
    \multirow{7}{*}{\makecell[l]{Reasoning}} 
     &MathVista$_{\text{mini}}$  & ${69.6}^{\phantom{{\uparrow0.0}}}$ & $\textbf{72.3}^{\textcolor{myorange}{\uparrow2.7}}$ & ${71.8}^{\textcolor{myorange}{\uparrow2.2}}$ & ${68.6}^{\phantom{\textcolor{myorange}{\uparrow0.0}}}$ & ${67.9}^{\phantom{\textcolor{myorange}{\uparrow0.0}}}$ & ${60.2}^{\phantom{\textcolor{myorange}{\uparrow0.0}}}$ & ${58.5}^{\phantom{\textcolor{myorange}{\uparrow0.0}}}$ \\
     &WeMath                     & ${61.5}^{\phantom{{\uparrow0.0}}}$ & $\textbf{69.4}^{\textcolor{myorange}{\uparrow7.9}}$ & ${60.8}^{\phantom{\textcolor{myorange}{\uparrow0.0}}}$ & ${61.3}^{\phantom{\textcolor{myorange}{\uparrow0.0}}}$ & ${62.0}^{\phantom{\textcolor{myorange}{\uparrow0.0}}}$ & ${45.1}^{\phantom{\textcolor{myorange}{\uparrow0.0}}}$ & ${44.1}^{\phantom{\textcolor{myorange}{\uparrow0.0}}}$ \\
     &MathVision                  & ${25.6}^{\phantom{{\uparrow0.0}}}$ & $\textbf{34.4}^{\textcolor{myorange}{\uparrow8.8}}$ & ${26.2}^{\textcolor{myorange}{\uparrow0.6}}$ & ${22.4}^{\phantom{\textcolor{myorange}{\uparrow0.0}}}$ & ${24.2}^{\phantom{\textcolor{myorange}{\uparrow0.0}}}$ & ${21.3}^{\phantom{\textcolor{myorange}{\uparrow0.0}}}$ & ${18.5}^{\phantom{\textcolor{myorange}{\uparrow0.0}}}$ \\
     &MMMU$_{\text{val}}$        & ${55.4}^{\phantom{{\uparrow0.0}}}$ & $\textbf{58.8}^{\textcolor{myorange}{\uparrow3.4}}$ & ${54.9}^{\phantom{\textcolor{myorange}{\uparrow0.0}}}$ & ${51.3}^{\phantom{\textcolor{myorange}{\uparrow0.0}}}$ & ${52.7}^{\phantom{\textcolor{myorange}{\uparrow0.0}}}$ & ${46.4}^{\phantom{\textcolor{myorange}{\uparrow0.0}}}$ & ${48.8}^{\phantom{\textcolor{myorange}{\uparrow0.0}}}$ \\
     &MMMU-Pro$_{\text{standard}}$ & ${37.4}^{\phantom{{\uparrow0.0}}}$ & $\textbf{39.9}^{\textcolor{myorange}{\uparrow2.5}}$ & ${38.0}^{\textcolor{myorange}{\uparrow0.6}}$ & ${36.3}^{\phantom{\textcolor{myorange}{\uparrow0.0}}}$ & ${35.3}^{\phantom{\textcolor{myorange}{\uparrow0.0}}}$ & ${31.1}^{\phantom{\textcolor{myorange}{\uparrow0.0}}}$ & ${28.0}^{\phantom{\textcolor{myorange}{\uparrow0.0}}}$ \\
     &MMMU-Pro$_{\text{vision}}$   & ${25.2}^{\phantom{{\uparrow0.0}}}$ & $\textbf{35.7}^{\textcolor{myorange}{\uparrow10.5}}$ & ${29.0}^{\textcolor{myorange}{\uparrow3.8}}$ & ${32.8}^{\phantom{\textcolor{myorange}{\uparrow0.0}}}$ & ${25.4}^{\phantom{\textcolor{myorange}{\uparrow0.0}}}$ & ${21.3}^{\phantom{\textcolor{myorange}{\uparrow0.0}}}$ & ${14.3}^{\phantom{\textcolor{myorange}{\uparrow0.0}}}$ \\
      \cmidrule{2-9} \rowcolor{gray!20} & Avg. & ${45.8}^{\phantom{\uparrow0.0}}$ & $\textbf{51.8}^{\textcolor{myorange}{\uparrow6.0}}$ & ${46.8}^{\textcolor{myorange}{\uparrow1.0}}$ & ${45.5}^{\phantom{\textcolor{myorange}{\uparrow0.0}}}$ & ${44.6}^{\phantom{\textcolor{myorange}{\uparrow0.0}}}$ & ${37.6}^{\phantom{\textcolor{myorange}{\uparrow0.0}}}$ & ${35.4}^{\phantom{\textcolor{myorange}{\uparrow0.0}}}$ \\
    \midrule
     \multirow{8}{*}{\makecell[l]{OCR \& Chart}} 
     &ChartQA                     & ${86.5}^{\phantom{{\uparrow0.0}}}$ & $\textbf{87.4}^{\textcolor{myorange}{\uparrow0.9}}$ & ${87.0}^{\textcolor{myorange}{\uparrow0.5}}$ & ${84.1}^{\phantom{\textcolor{myorange}{\uparrow0.0}}}$ & ${87.1}^{\phantom{\textcolor{myorange}{\uparrow0.0}}}$ & ${83.4}^{\phantom{\textcolor{myorange}{\uparrow0.0}}}$ & ${80.0}^{\phantom{\textcolor{myorange}{\uparrow0.0}}}$ \\
     &CharXiv$_{\text{DQ}}$       & ${70.9}^{\phantom{{\uparrow0.0}}}$ & ${68.4}^{\phantom{{\uparrow0.1}}}$ & $\textbf{71.2}^{\textcolor{myorange}{\uparrow0.3}}$ & ${69.8}^{\phantom{\textcolor{myorange}{\uparrow0.0}}}$ & ${63.8}^{\phantom{\textcolor{myorange}{\uparrow0.0}}}$ & ${58.2}^{\phantom{\textcolor{myorange}{\uparrow0.0}}}$ & ${47.6}^{\phantom{\textcolor{myorange}{\uparrow0.0}}}$ \\
     &DocVQA                      & ${95.0}^{\phantom{{\uparrow0.0}}}$ & ${91.9}^{\phantom{\textcolor{myorange}{\uparrow0.0}}}$ & $\textbf{95.0}^{\textcolor{myorange}{\uparrow0.0}}$ & ${94.9}^{\phantom{\textcolor{myorange}{\uparrow0.0}}}$ & ${94.4}^{\phantom{\textcolor{myorange}{\uparrow0.0}}}$ & ${92.7}^{\phantom{\textcolor{myorange}{\uparrow0.0}}}$ & ${87.2}^{\phantom{\textcolor{myorange}{\uparrow0.0}}}$ \\
     &OCRBench                    & ${82.9}^{\phantom{{\uparrow0.0}}}$ & ${81.7}^{\phantom{\textcolor{myorange}{\uparrow0.0}}}$ & ${82.3}^{\phantom{\textcolor{myorange}{\uparrow0.0}}}$ & $\textbf{84.2}^{\phantom{\textcolor{myorange}{\uparrow0.0}}}$ & ${80.0}^{\phantom{\textcolor{myorange}{\uparrow0.0}}}$ & ${79.2}^{\phantom{\textcolor{myorange}{\uparrow0.0}}}$ & ${62.1}^{\phantom{\textcolor{myorange}{\uparrow0.0}}}$ \\
     &AI2D$_{\text{w M}}$          & ${84.2}^{\phantom{{\uparrow0.0}}}$ & ${83.7}^{\phantom{\textcolor{myorange}{\uparrow0.0}}}$ & $\textbf{84.3}^{\textcolor{myorange}{\uparrow0.1}}$ & ${82.6}^{\phantom{\textcolor{myorange}{\uparrow0.0}}}$ & ${83.6}^{\phantom{\textcolor{myorange}{\uparrow0.0}}}$ & ${78.6}^{\phantom{\textcolor{myorange}{\uparrow0.0}}}$ & ${81.4}^{\phantom{\textcolor{myorange}{\uparrow0.0}}}$ \\
     &AI2D$_{\text{w/o M}}$              & $\textbf{94.1}^{\phantom{{\uparrow0.0}}}$ & ${93.7}^{\phantom{\textcolor{myorange}{\uparrow0.0}}}$ & ${93.9}^{\phantom{\textcolor{myorange}{\uparrow0.0}}}$ & ${93.4}^{\phantom{\textcolor{myorange}{\uparrow0.0}}}$ & ${93.3}^{\phantom{\textcolor{myorange}{\uparrow0.0}}}$ & ${90.7}^{\phantom{\textcolor{myorange}{\uparrow0.0}}}$ & ${90.8}^{\phantom{\textcolor{myorange}{\uparrow0.0}}}$ \\
     &InfoVQA               & ${78.4}^{\phantom{{\uparrow0.0}}}$ & ${76.6}^{\phantom{\textcolor{myorange}{\uparrow0.0}}}$ & ${78.7}^{\textcolor{myorange}{\uparrow0.3}}$ & $\textbf{81.7}^{\phantom{\textcolor{myorange}{\uparrow0.0}}}$ & ${76.1}^{\phantom{\textcolor{myorange}{\uparrow0.0}}}$ & ${75.6}^{\phantom{\textcolor{myorange}{\uparrow0.0}}}$ & ${68.8}^{\phantom{\textcolor{myorange}{\uparrow0.0}}}$ \\
     
    \cmidrule{2-9} \rowcolor{gray!20} & Avg. & ${84.6}^{\phantom{\uparrow0.0}}$ & ${83.3}^{\phantom{\textcolor{myorange}{\uparrow0.0}}}$ & $\textbf{84.6}^{{\textcolor{myorange}{\uparrow0.0}}}$ & ${84.4}^{\phantom{\textcolor{myorange}{\uparrow0.0}}}$ & ${82.6}^{\phantom{\textcolor{myorange}{\uparrow0.0}}}$ & ${79.8}^{\phantom{\textcolor{myorange}{\uparrow0.0}}}$ & ${74.0}^{\phantom{\textcolor{myorange}{\uparrow0.0}}}$ \\
    \midrule
     \multirow{5}{*}{\makecell[l]{Others}} 
     &PixmoCount                  & ${62.2}^{\phantom{{\uparrow0.0}}}$ & ${65.7}^{\textcolor{myorange}{\uparrow3.5}}$ & $\textbf{71.1}^{\textcolor{myorange}{\uparrow8.9}}$ & $\textbf{63.3}^{\phantom{\textcolor{myorange}{\uparrow0.0}}}$ & ${52.2}^{\phantom{\textcolor{myorange}{\uparrow0.0}}}$ & ${50.9}^{\phantom{\textcolor{myorange}{\uparrow0.0}}}$ & ${49.3}^{\phantom{\textcolor{myorange}{\uparrow0.0}}}$ \\
     &CountBench                  & ${88.2}^{\phantom{{\uparrow0.0}}}$ & ${86.8}^{\phantom{\textcolor{myorange}{\uparrow0.0}}}$ & $\textbf{88.6}^{\textcolor{myorange}{\uparrow0.4}}$ & ${86.4}^{\phantom{\textcolor{myorange}{\uparrow0.0}}}$ & ${79.8}^{\phantom{\textcolor{myorange}{\uparrow0.0}}}$ & ${72.5}^{\phantom{\textcolor{myorange}{\uparrow0.0}}}$ & ${78.4}^{\phantom{\textcolor{myorange}{\uparrow0.0}}}$ \\ 
     &VL-RewardBench              & ${47.7}^{\phantom{{\uparrow0.0}}}$ & ${44.0}^{\phantom{\textcolor{myorange}{\uparrow0.0}}}$ & $\textbf{49.7}^{\textcolor{myorange}{\uparrow2.0}}$ & ${49.7}^{\phantom{\textcolor{myorange}{\uparrow0.0}}}$ & ${48.2}^{\phantom{\textcolor{myorange}{\uparrow0.0}}}$ & ${42.1}^{\phantom{\textcolor{myorange}{\uparrow0.0}}}$ & ${44.5}^{\phantom{\textcolor{myorange}{\uparrow0.0}}}$ \\
     &V\text{*}                          & ${78.0}^{\phantom{{\uparrow0.0}}}$ & $\textbf{79.1}^{\textcolor{myorange}{\uparrow1.1}}$ & ${78.0}^{\textcolor{myorange}{\uparrow0.0}}$ & ${77.0}^{\phantom{\textcolor{myorange}{\uparrow0.0}}}$ & ${74.9}^{\phantom{\textcolor{myorange}{\uparrow0.0}}}$ & ${69.6}^{\phantom{\textcolor{myorange}{\uparrow0.0}}}$ & ${72.3}^{\phantom{\textcolor{myorange}{\uparrow0.0}}}$ \\
     \cmidrule{2-9} \rowcolor{gray!20} & Avg. & ${69.0}^{\phantom{{\uparrow0.0}}}$ & ${66.0}^{\phantom{{\uparrow0.0}}}$ & $\textbf{71.6}^{\textcolor{myorange}{\uparrow{2.6}}}$ & ${69.1}^{\phantom{\textcolor{myorange}{\uparrow0.0}}}$ & ${63.8}^{\phantom{\textcolor{myorange}{\uparrow0.0}}}$ & ${58.8}^{\phantom{\textcolor{myorange}{\uparrow0.0}}}$ & ${61.1}^{\phantom{\textcolor{myorange}{\uparrow0.0}}}$ \\
    \bottomrule
    \end{tabular}
    \vspace{-2mm}
    \end{threeparttable}
  }
  \end{table*}

\subsection{Infrastructure}

\begin{figure}[t]
    \centering
    \includegraphics[width=\textwidth]{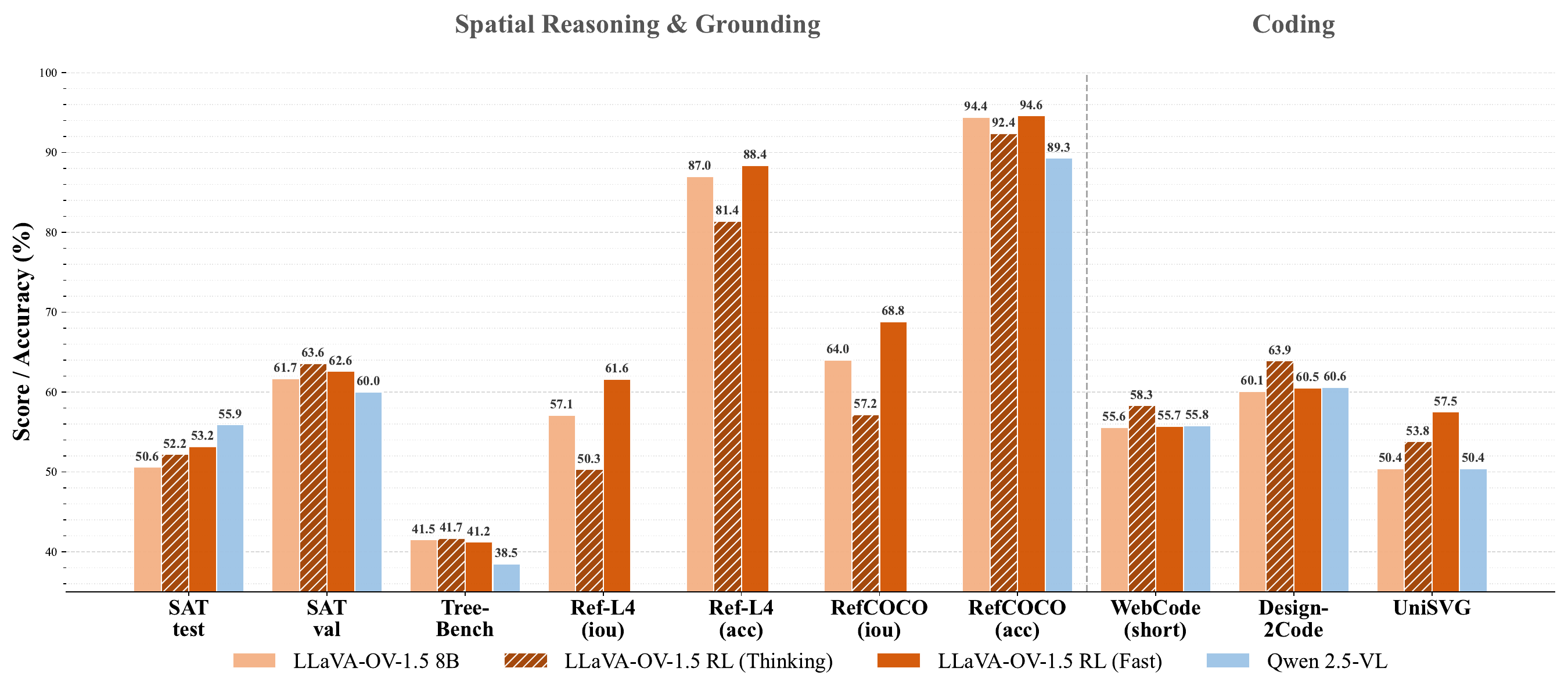}
    \vspace{-5mm}
    \caption{Performance comparison of LLaVA-OV-1.5 and corresponding RL version on Spatial Reasoning \& Grounding and Coding tasks. }
    \label{fig:rl_benchmark}
    \vspace{-3mm}
\end{figure}

\textbf{\textit{Load Balancing via Data Packing.}} A major source of training inefficiency arises from padding, where batch samples are standardized by adding padding tokens. This results in significant computational overhead and poor GPU utilization, particularly with heterogeneous multimodal data. To mitigate this, we propose an offline parallel data packing method that consolidates multiple shorter samples into packed sequences during preprocessing. Our approach employs hash buckets to handle large-scale data efficiently and leverages multi-threaded, strategy-aware batching to control packing success rate, sample count, and batch composition. Unlike online packing, which operates dynamically at runtime, our method processes entire datasets or large contiguous chunks offline, ensuring uniform output lengths. This yields up to an $11\times$ compression ratio on 85 million pretraining samples, substantially improving efficiency.

\textbf{\textit{Hybrid Parallelism Framework.}} 
We adopt AIAK-Training-LLM\footnote{AIAK-Training-LLM: Baidu Cloud’s optimized Megatron-LM} built upon Megatron-LM~\citep{megatron-lm} as our training framework. Its transformer engine and specialized optimizations enable efficient mid-training of \vlname-8B with a context length of 8K. By leveraging distributed optimizer parallelism and uniform recomputation, the mid-training process is conducted at native resolution on 85 million captions using 128 $\times$ A800 GPUs over 3.7 days.
\section{Post-training}
\label{sec:posttraining}

To further enhance \vlname's performance on multimodal reasoning tasks, we perform a reinforcement learning-based post-training stage on top of the supervised \vlname-Instruct model. 

\subsection{RL Training Data}

\paragraph{Discrepancy-Driven Data Selection.} We curate the training data by measuring the divergence between \textit{Pass@N} and \textit{Pass@1} performance on diverse benchmarks. A significant gap indicates that the model possesses the \textit{latent capability} to solve the task, as correct solutions do appear within its sampling distribution, yet its policy distribution fails to reliably assign high probability to the correct reasoning path. Under this lens, RL serves as an \textbf{elicitation} mechanism rather than knowledge injection, redirecting probability mass toward solutions the model can already generate but does not consistently prioritize.
This selection paradigm ensures high training efficiency by targeting the model's effective learnable boundary, avoiding trivial tasks it has already mastered or unsolvable ones beyond its current grasp. 

We construct the RL training corpus by aggregating diverse public data sources, including ViRL~\citep{yuan2024virl39k}, WebCode2M~\citep{zhou2024webcode2m}, UniSVG~\citep{li2025unisvg}, Ref-L4~\citep{chen2024revisiting}, VigoRL-SA~\citep{sarch2025vigorl}, VigoRL-SAT~\citep{sarch2025vigorl}, PixmoCount~\citep{molmo}, AI2D~\citep{kembhavi2016diagram}, and InfoVQA~\citep{mathew2022infographicvqa}. These sources cover a wide range of capabilities such as STEM reasoning, coding, grounding, counting, spatial reasoning, diagram understanding, and OCR.

\paragraph{Reward-Based Sampling.} To further filter the high-quality training instances, we employ a \textbf{reward-based sampling} strategy. Specifically, we generate multiple candidate responses for each sample using the base model and compute their automatic rewards. We then retain only those examples where the average reward across candidates falls within a specified range. This filtering process effectively discards both trivial and unsolvable cases, biasing the corpus toward medium-difficulty instances that provide the most valuable learning signal.

Finally, we obtain a unified RL corpus of about 67K instances. The detailed composition is shown in Figure~\ref{fig:rl_stage_distribution} (a). Specifically, \textbf{STEM} data (38.9K) comes from ViRL39K; \textbf{Grounding} (15K) aggregates Ref-L4 and VigoRL-SA; \textbf{Spatial} (4.2K) and \textbf{Counting} (2.8K) tasks are sourced from VigoRL-SAT and PixmoCount; \textbf{Coding} (4K) combines WebCode2M and UniSVG; while \textbf{OCR} (2K) and \textbf{Diagram} (0.2K) samples are selected from InfoVQA and AI2D, respectively. Each instance additionally records whether it is prompted in a short answer-only style or a longer chain-of-thought style, which we later exploit when designing different RL curricula over the same underlying pool.

\subsection{Reward System}

Our RL setup employs a rule-based reward paradigm, where rewards are derived directly from task outcomes rather than learned preference models. Since different answer types require distinct verification strategies, we design \textbf{answer-type-specific scoring rules}.

For \textbf{STEM} questions from ViRL39K, we face a key challenge: the model may express the same correct answer in vastly different formats. To address this, we implement a multi-stage verification pipeline. First, we extract answers via flexible parsing (prioritizing structured tags like \texttt{<answer>} but falling back to heuristics such as ``Final Answer:'' when necessary). Second, we normalize LaTeX artifacts (e.g., unifying \texttt{$\backslash$frac12} and \texttt{$\backslash$frac\{1\}\{2\}}). Finally, for numerical reasoning, we perform symbolic equivalence checking rather than string matching—thus $\frac{1}{2}$, $0.5$, and even $\frac{2}{4}$ are all recognized as correct if the ground truth is any one of them. For multiple-choice problems, we match extracted option labels (A/B/C/D) against the reference, tolerating common formatting variations like \texttt{(A)}, \texttt{**A**}, or \texttt{A.}. In \textbf{Coding} benchmarks, WebCode2M samples are rewarded based on token- and tag-level overlap with the reference code, while UniSVG further incorporates an SVG rendering similarity score in $[0,1]$ to encourage perceptually matched graphics.

\textbf{Grounding} data from Ref-L4 and VigoRL-SA are evaluated by the intersection-over-union (IoU) between predicted and reference bounding boxes, combined with standard accuracy for associated multiple-choice queries. \textbf{Spatial-reasoning} problems from VigoRL-SAT are scored purely by answer accuracy. For \textbf{Counting} tasks such as PixmoCount, we extract the final numeric token and require exact equality with the gold count.
\textbf{OCR} instances from InfoVQA use text-similarity based rewards between the predicted and reference strings, and \textbf{Diagram} questions from AI2D are judged by multiple-choice accuracy. These answer-type-aware rules are collapsed into a single scalar reward per candidate response. 

\subsection{Training Procedure}

We employ Group Relative Policy Optimization (GRPO)~\citep{shao2024deepseekmath} as our core reinforcement learning algorithm. To maximize training efficiency and throughput, we adopt the GRPO implementation in {AReaL}~\citep{fu2025areal}, a state-of-the-art asynchronous RL framework. AReaL decouples generation from training, allowing rollout workers to continuously generate data while trainer workers update the model in parallel, significantly improving GPU utilization compared to synchronous implementations.

Regarding the optimization objective, we simplify the standard GRPO formulation by \textbf{omitting the KL divergence penalty}, relying instead on PPO-style clipping to maintain training stability. Furthermore, we discard the explicit format reward commonly used to enforce structural constraints (e.g., XML tagging). Instead, we rely solely on outcome-based correctness rewards. 
To better exploit the structure of our RL corpus, we employ a two-stage curriculum, as shown in Figure~\ref{fig:rl_stage_distribution}(b) and (c):

\begin{enumerate}[leftmargin=*]
    \item \textbf{Stage~1: Answer-only RL on normal data.} In the first stage, we train exclusively on the normal split, where instructions ask the model to output only the final answer. For these tasks we use the prompt
    \begin{quote}
    \small
    \texttt{Put ONLY your final answer within <answer></answer>.}
    \end{quote}
    This warm-up stage solidifies basic perceptual skills like counting, serving as a critical foundation for subsequent reasoning tasks. This curriculum ensures the model retains precision on simple problems and avoids ``over-thinking'' when later advancing to complex reasoning chains.

    \item \textbf{Stage~2: Chain-of-thought RL on long-reasoning data.} In the second stage, we switch to the long-reasoning split and encourage the model to produce explicit reasoning traces. The instruction for these tasks is
    \begin{quote}
    \small
    \texttt{Think and solve the following question step by step. Please put your thinking and analysis procedure within <think></think>. Put ONLY your final answer within <answer></answer>.}
    \end{quote}
    The reward is still computed only from the content within \texttt{<answer></answer>}, ensuring that the optimization target remains answer correctness while the reasoning tokens inside \texttt{<think></think>} serve as auxiliary guidance.
\end{enumerate}

A naive second stage that uses only long-reasoning tasks can cause the model to forget short, perception-heavy skills. To mitigate this, we interleave a small proportion of normal-set examples into Stage~2 mini-batches. These samples continue to use the answer-only prompt and reward, acting as an anchor that preserves the model's competence on concise tasks while RL emphasizes deeper reasoning. Overall, this two-stage, mixed-prompt curriculum allows \vlname-RL to simultaneously strengthen long-horizon reasoning and maintain strong performance on standard vision-language benchmarks, details in Section~\ref{subsec:rl_evaluation}.

\section{Experiments}

\subsection{Overall Performance}
We use LMMs-Eval~\cite{lmms_eval} with the default prompt to evaluate the performance of \vlname\ across multiple benchmarks in four categories of downstream tasks:(1) General Visual Question Answering (VQA): MMStar~\citep{mmstar}, MMEBench series~\citep{mme}, MME-RealWorld series~\citep{mmerealworld}, SeedBench~\citep{seedbench}, Seed-Bench-2-Plus~\citep{seedbench2}, CV-Bench~\citep{tong2024cambrian}, and RealWorldQA~\citep{realworldqa}. (2) Multimodal Reasoning: MathVista~\citep{mathvista}, WeMath~\citep{wemath}, MathVision~\citep{mathvision}, MMMU~\citep{mmmu}, and MMMU-Pro series~\citep{mmmupro}.(3) OCR \& Chart Understanding: ChartQA~\citep{chartqa}, CharXiv~\citep{charxiv}, DocVQA~\citep{docvqa}, OCRBench~\citep{ocrbench}, AI2D~\citep{ai2d}, and InfoVQA~\citep{infovqa}.(4) Others: PixmoCount~\citep{molmo}, CountBench~\citep{paiss2023teaching}, VL-RewardBench~\citep{li2025vl}, and V$^{*}$~\citep{wu2024v}. As shown in Tab.~\ref{tab:main}, \vlname-8B surpasses Qwen2.5-VL-7B on 18 of 27 benchmarks and \vlname-4B surpasses Qwen2.5-VL-3B on 27 of 27 benchmarks.

\subsection{General Visual Question Answering}
As detailed in Tab.~\ref{tab:main}, we evaluate the general visual question answering capability of \vlname\ across multiple benchmarks, and \vlname-8B demonstrates superior performance on MMStar (67.7), MMBench$_{\text{en}}$ (84.1), MME-RealWorld$_{\text{en}}$ (62.3), MME-RealWorld$_{\text{cn}}$ (56.1), CV-Bench (80.8), and ScienceQA (95.0). Besides, \vlname\ also presents comparable performance on MMBench$_{\text{cn}}$ (81.0), SeedBench$_{\text{image}}$ (77.3), SEED-Bench-2-Plus (69.2), and RealWorldQA (68.1).

\definecolor{RowLightBlue}{HTML}{E8F4FF}

\begin{table*}[!t]
    \centering
    \fontsize{6.6pt}{5.5pt}\selectfont
    \setlength\tabcolsep{2.2pt}
    \renewcommand{\arraystretch}{1.5}
    \caption{Comparison of RICE-ViT with other vision encoders using the LLaVA-NeXT framework. All models are evaluated using identical configurations: Qwen2.5-7B as the language model, LLaVA-NeXT training data, and the same training pipeline. To ensure fair comparison, we adopt LLaVA-NeXT's tiling strategy (up to 2×2+1 tiles) for handling high-resolution images, as many vision encoders do not support native resolution processing. }
    \label{tab:comparison_in_llava}
    \begin{tabular}{ll|*{7}{c}|c|*{7}{c}|c}
        \toprule
        \multicolumn{2}{c|}{\textbf{Model Configuration}} & 
        \multicolumn{8}{c|}{\textbf{OCR \& Document Understanding}} & 
        \multicolumn{8}{c}{\textbf{General Vision Understanding}} \\
        \midrule
        \textbf{Method} & \textbf{Vision Tower} & 
        \rotatebox{90}{InfoVQA} & 
        \rotatebox{90}{DocVQA} & 
        \rotatebox{90}{ChartQA} & 
        \rotatebox{90}{TextVQA} & 
        \rotatebox{90}{OCRBench} & 
        \rotatebox{90}{OCRBenchV2} & 
        \rotatebox{90}{LiveXivVQA} & 
        \rotatebox{90}{\textbf{OCR Avg}} & 
        \rotatebox{90}{AI2D} & 
        \rotatebox{90}{MMB$^\text{EN}$} & 
        \rotatebox{90}{MME$^\text{Cog}$} & 
        \rotatebox{90}{MME$^\text{Per}$} & 
        \rotatebox{90}{POPE} & 
        \rotatebox{90}{RealworldQA} & 
        \rotatebox{90}{MMStar} & 
        \rotatebox{90}{\textbf{Other Avg}} \\
        \midrule

        CLIP & ViT-L-14-336px & 38.9 & 75.2 & 66.5 & 62.5 & 52.5 & 23.0 & 47.4 & 52.3 & 73.2 & 74.6 & 48.0 & 75.6 & \textbf{88.8} & \textbf{63.7} & 49.0 & 67.6 \\
        MLCD & ViT-L-14-336px & 43.5 & 76.5 & 67.8 & 61.7 & 53.1 & 24.0 & 48.4 & 53.6 & 77.0 & 76.4 & 54.1 & 79.9 & 88.7 & 61.1 & 51.0 & 69.7 \\
        AIMv2 & ViT-L-14-336px & 35.4 & 77.2 & \textbf{72.7} & \textbf{65.9} & 57.2 & 23.9 & 47.3 & 54.2 & 75.4 & \textbf{78.6} & 48.3 & 75.0 & 88.4 & 62.2 & 50.2 & 68.3 \\
        \rowcolor[rgb]{0.992,0.953,0.906} RICE-ViT & ViT-L-14-336px & \textbf{45.2} & \textbf{79.2} & 72.3 & \textbf{65.9} & \textbf{57.5} & \textbf{24.1} & \textbf{48.9} & \textbf{56.2} & \textbf{77.9} & 76.6 & \textbf{54.6} & \textbf{80.7} & 88.5 & 63.1 & \textbf{51.8} & \textbf{70.5} \\
        \midrule
        DFN5B & ViT-H-14-378px & 38.6 & 70.9 & 64.4 & 59.4 & 47.3 & 21.9 & 46.2 & 49.8 & 73.5 & 73.4 & 45.8 & 76.9 & 88.6 & 59.9 & 49.1 & 66.7 \\
        SigLIP & ViT-SO400M-14-384px & 41.4 & 76.7 & 69.3 & 64.7 & 55.4 & 24.0 & 48.4 & 54.3 & 76.2 & 77.0 & 46.1 & 79.9 & 88.8 & \textbf{63.7} & 47.3 & 68.4 \\
        SigLIPv2 & ViT-SO400M-14-384px & 43.7 & 79.1 & 70.2 & 66.2 & 58.7 & 25.4 & 48.6 & 56.0 & \textbf{77.0} & 77.1 & 46.6 & \textbf{80.4} & \textbf{89.3} & 63.4 & \textbf{52.8} & 69.5 \\
        \rowcolor[rgb]{0.992,0.953,0.906} RICE-ViT & ViT-L-14-378px & \textbf{48.1} & \textbf{82.6} & \textbf{75.1} & \textbf{66.2} & \textbf{58.8} & \textbf{25.8} & \textbf{49.5} & \textbf{58.0} & 76.5 & \textbf{77.6} & \textbf{54.1} & 79.0 & 89.1 & 62.9 & 51.2 & \textbf{70.1} \\
        \midrule
        SigLIPv2 & ViT-SO400M-16-560px & 50.2 & 86.2 & 77.4 & \textbf{70.2} & \textbf{62.7} & \textbf{26.5} & 52.9 & 60.9 & \textbf{77.0} & 76.5 & 53.5 & \textbf{79.9} & \textbf{89.3} & \textbf{68.2} & \textbf{53.1} & \textbf{71.1} \\
        \rowcolor[rgb]{0.992,0.953,0.906} RICE-ViT & ViT-L-14-560px & \textbf{53.2}& \textbf{87.4} & \textbf{78.1} & 69.0 & 60.7 & 26.1 & \textbf{53.0} & \textbf{61.1} & 76.9 & \textbf{78.6} & \textbf{56.3} & 79.3 & 88.9 & 65.1 & 50.5 & 70.8 \\
         Qwen-ViT from Qwen2.5-VL 7B & ViT-H-14-560px & \textbf{55.9} & 85.8 & 78.8 & 73.7 & 66.2 & 26.8 & 53.4 & 62.9 & 78.8 & 78.4 & \textbf{62.0} & 80.8 & 88.6 & 64.2 & 55.0 & 72.5 \\
        \rowcolor[rgb]{0.992,0.953,0.906}
        RICE-ViT from OV-1.5 3B & ViT-L-14-560px & 53.7 & \textbf{87.1} & \textbf{81.9} & \textbf{73.8} & \textbf{73.3} & \textbf{30.4} & \textbf{53.6} & \textbf{64.8} & \textbf{80.3} & \textbf{79.6} & 58.6 & \textbf{82.2} & \textbf{89.0} & \textbf{67.3} & \textbf{56.6} & \textbf{73.4} \\
        \bottomrule
    \end{tabular}
    
\end{table*}

\subsection{Multimodal Reasoning}
\vlname\ exhibits superior multimodal reasoning capabilities compared to Qwen2.5-VL. Specifically, \vlname-4B outperforms Qwen2.5-VL-3B on all evaluated benchmarks, leading in MathVista$_{\text{mini}}$ (67.9), WeMath (24.9), MathVision (24.2), MMMU$_{\text{val}}$ (52.7), MMMU-Pro$_{\text{standard}}$ (35.3), and MMMU-Pro$_{\text{vision}}$ (25.4). Notably, \vlname-4B surpasses LLaVA-OneVision-7B across all benchmarks. Compared with Qwen2.5-VL-7B, \vlname-8B also demonstrates gains of 1.0\%, 0.3\%, 3.2\%, 4.1\%, and 1.1\% on MatchVista$_{\text{mini}}$, WeMath, MathVision, MMMU$_{\text{val}}$, and MMMU-Pro$_{\text{standard}}$.

\subsection{OCR \& Chart Understanding}
The interpretation of visual data, including documents and charts, requires a sophisticated array of skills from multimodal large language models, ranging from low-level Optical Character Recognition (OCR) to high-level semantic reasoning. To thoroughly evaluate these capabilities, we assess \vlname\ across seven challenging benchmarks. \vlname-8B demonstrates robust outcomes on ChartQA (86.5), CharXiv$_{\text{DQ}}$ (74.1), DocVQA (95.0), AI2D$_{\text{w M}}$ (84.2), and AI2D$_{\text{w/o M}}$ (94.1). Notably, \vlname-4B outperforms Qwen2.5-VL-3B on all seven benchmarks.

\subsection{Others}
To further elucidate the capabilities of \vlname, we extend our evaluation to include PixmoCount, CountBench, VL-RewardBench, and V$^*$. \vlname-8B records scores of 62.2 on PixmoCount, 88.2 on CountBench, 46.7 on VL-RewardBench, and 78.0 on V$^*$, demonstrating proficiency in counting, visual perception, and visual grounding, on par with Qwen2.5-VL-7B.

\subsection{Performance of RL Post-training}
\label{subsec:rl_evaluation}
To validate the effectiveness of our lightweight RL post-training, we compare the RL-enhanced models (LLaVA-OV-1.5 RL-8B) against their supervised baselines and the strong competitor Qwen2.5-VL-7B. We report results for both standard "fast" inference and "thinking" mode (where the model generates reasoning chains) as shown in Tab.~\ref{tab:main}.

\textbf{Core Capability Enhancement.} As presented in Tab.~\ref{tab:main}, RL post-training yields consistent gains across major benchmarks.
 (1) \textbf{Multimodal Reasoning}: The most substantial improvements are observed in multimodal reasoning tasks. In "thinking" mode, our model achieves dramatic gains on WeMath (+7.9), MathVision (+8.8), and MMMU-Pro$_{\text{vision}}$ (+10.5), demonstrating that the RL-induced chain-of-thought capability effectively unlocks deeper problem-solving skills.
 (2) \textbf{General VQA \& OCR}: On standard benchmarks like MMBench and DocVQA, RL maintains or slightly improves the already strong SFT performance, ensuring no regression in general capabilities while specializing in hard reasoning.

\textbf{Extended Capability Analysis.} Beyond the core benchmarks, we utilize Figure~\ref{fig:rl_benchmark} to analyze the impact of RL on specific vertical capabilities not fully covered in Tab.~\ref{tab:main}.
(1) \textbf{Spatial Reasoning \& Grounding}: As shown in Figure~\ref{fig:rl_benchmark}, RL significantly enhances fine-grained perception. The RL model (Fast mode) consistently outperforms the SFT baseline on spatial tasks like SAT and Ref-L4. Interestingly, while "thinking" mode aids reasoning, it sometimes yields lower scores than "fast" mode on strictly perceptual metrics (e.g., Ref-L4 IoU), suggesting that verbose generation may occasionally interfere with precise coordinate regression.
(2) \textbf{Coding}: In the coding domain (WebCode, UniSVG), Figure~\ref{fig:rl_benchmark} shows that our RL models achieve consistent gains. The "thinking" mode proves particularly effective here, achieving the highest scores on Design2Code and UniSVG, indicating that chain-of-thought reasoning is beneficial for structural code generation.

These results confirm that our rule-based RL framework effectively enhances both the reasoning depth (via thinking chains) and perceptual precision (via direct feedback) of \vlname, positioning it as a state-of-the-art open model in the 8B parameter class. 

\begin{figure}[t!]
  \centering
  \includegraphics[width=1.0\linewidth]{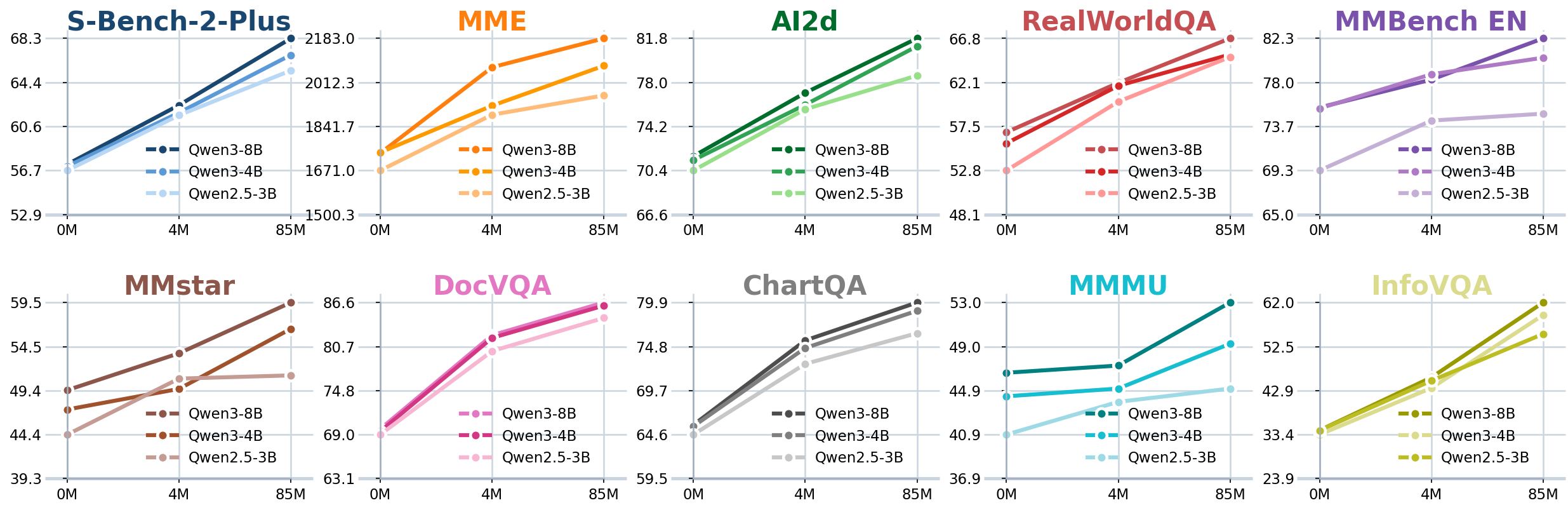}
  \vspace{-7mm}
  \caption{Performance comparison across different mid-training data scales on various benchmarks. Models initially undergo pre-training on LLaVA-558K and are then subjected to mid-training at different data scales (4M, 85M), followed by fine-tuning using the LLaVA-NeXT~\citep{llavanext} SFT framework. 0M denotes native pre-training without the mid-training stage.}
  \vspace{-2mm}
  \label{scale}
\end{figure}

\begin{figure}[t!]
  \centering
      \includegraphics[width=0.85\linewidth]{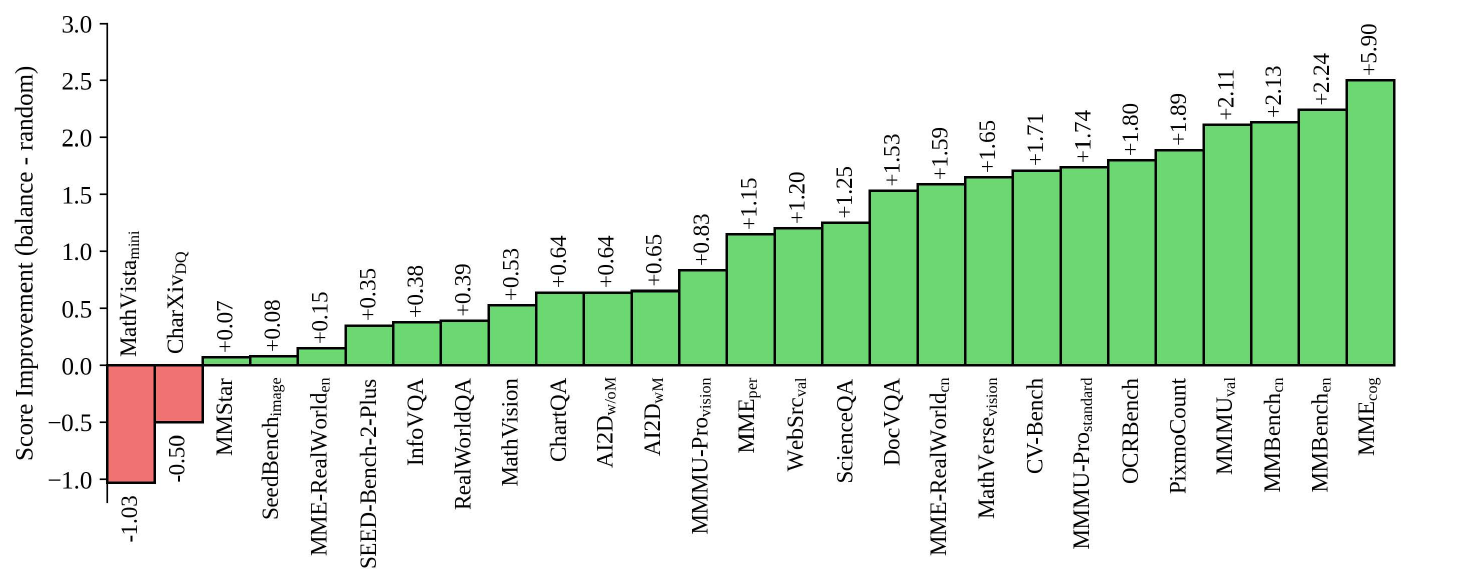}
      \vspace{-3mm}
  \caption{Experimental results using 2M blanced and unbalanced mid-training samples (LLaVA-NeXT-780k as the SFT data) show that using a balanced mid-training dataset yields consistent improvements over a random sampling strategy.}
  \label{balance}
\end{figure}

\subsection{Ablation Study}
\subsubsection{Comparison of Different Vision Encoders}
In Tab.~\ref{tab:comparison_in_llava}, we evaluate various vision encoders, CLIP~\citep{clip}, MLCD~\citep{mlcd}, AIMv2~\citep{fini2025multimodal}, DFN~\citep{dfn}, SigLIP~\citep{siglip}, SigLIPv2~\citep{siglipv2}, and RICE-ViT~\citep{rice}, within the LLaVA-NeXT framework. At a resolution of 336 pixels, RICE-ViT surpasses CLIP across all benchmarks, achieving significant gains in InfoVQA (+6.3\%) and OCRBench (+5.0\%). It also demonstrates significant improvements in document understanding compared to AIMv2. At 378 pixels, RICE-ViT outperforms computationally intensive models such as SigLIPv2 in 9 of 14 benchmarks, notably in InfoVQA (+4.4\%), DocVQA (+3.5\%), and ChartQA (+4.9\%). These results position RICE-ViT as a leading vision encoder, enhancing OCR capabilities and providing robust visual understanding, crucial for applications requiring advanced document analysis and visual reasoning. In addition, we further compare the performance of RICE-ViT with that of Qwen-ViT after incorporating LMM training, where RICE-ViT is derived from LLaVA-OneVision-1.5-3B and Qwen-ViT is derived from Qwen2.5-VL-7B. In the areas of OCR \& Document Understanding and General Vision Understanding, RICE-ViT demonstrates average performance improvements of 1.9\% and 0.9\% compared to Qwen-ViT.

\begin{figure}[t!]
  \centering
  \includegraphics[width=0.95\linewidth]{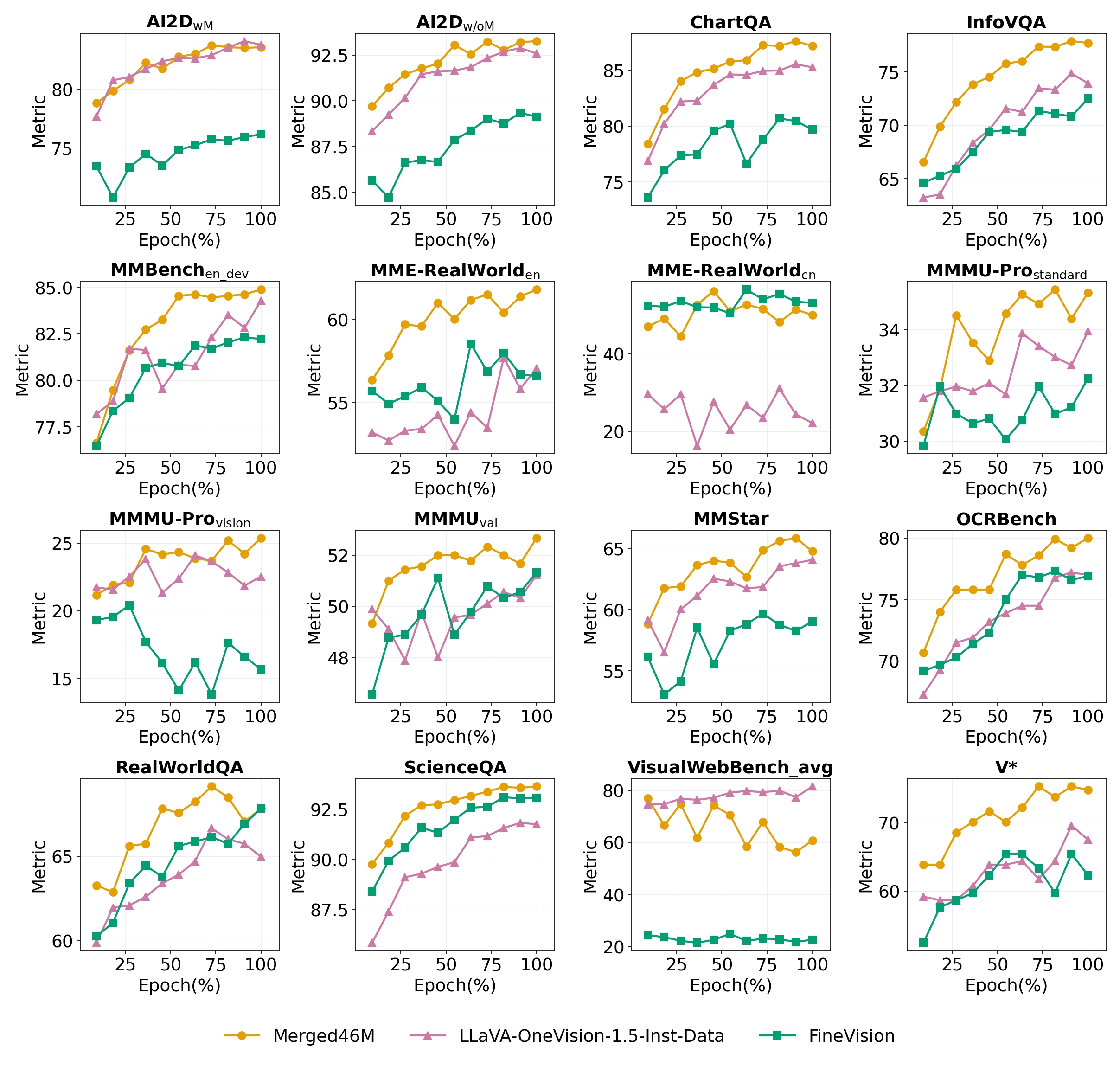}
  \vspace{-3mm}
  \caption{Performance comparison of three datasets (Merge46M, FineVision, and LLaVA-OneVision-1.5-Inst-Data) across 16 benchmarks during the SFT phase, demonstrating the superiority of Merge46M on most benchmarks.}
  \label{data_compare}
  \vspace{-3mm}
\end{figure}

\subsubsection{Mid-Training Data Scaling}
As depicted in Fig.~\ref{scale}, we present the performance of three different LMMs trained with various scales of mid-training data across ten distinct benchmarks. Employing LLaVA-558K for language-image alignment and standard LLaVA-Next instruction tuning, our findings indicate that scaling the data volume during the high-quality knowledge learning phase consistently enhances model performance across all benchmarks. These results not only underscore the high quality and scalability of the proposed \pretraindb\ dataset but also confirm the efficacy of data scaling in improving the performance of LMMs.

\subsubsection{Effectiveness of Concept Balance}
Fig.~\ref{fig:5a} illustrates that after implementing concept balancing, \pretraindb\ exhibits a smoother distribution, thereby enhancing the model's capability to assimilate a more comprehensive set of knowledge. To further validate the effect of concept balance, we conducted a comparative analysis of models trained on 2M concept-balanced data versus those trained on 2M data obtained through random sampling. As indicated in Fig.~\ref{balance}, the concept-balanced 2M data set demonstrates superior performance in 25 of 27 evaluated downstream benchmarks.

\subsubsection{Instruction Data Quality and Scaling}
To enhance performance across diverse VQA tasks, we compile 124 types of instruction data (LLaVA-OneVision-1.5-Inst-Data) for SFT training. We further scale the model capabilities by deduplicating and merging the recently proposed FineVision dataset~\citep{finevision2025}, resulting in the Merged46M SFT dataset. To maintain consistent training steps, we double the batch size for Merged46M due to its larger scale. Fig.~\ref{data_compare} shows performance comparisons on 16 benchmarks during SFT using three datasets: LLaVA-OneVision-1.5-Inst-Data, FineVision, and Merged46M. LLaVA-OneVision-1.5-Inst-Data achieves performance comparable to FineVision, while the Merged46M dataset delivers the best results across nearly all benchmarks. 

\section{Conclusions}
In this work, we introduce \vlname, a family of large multimodal models that establishes a new paradigm for constructing high-performance vision-language systems with improved efficiency and reproducibility. We demonstrate the feasibility of training competitive multimodal models from scratch under strict constraints. Our contributions are threefold: a large-scale, curated multimodal dataset; an efficient end-to-end training framework operable under a limited budget; and extensive empirical results demonstrating state-of-the-art performance across diverse benchmarks. The model excels particularly in resource-constrained settings, surpassing strong baselines such as Qwen2.5-VL-7B. This study underscores how open and efficient frameworks can drive progress in multimodal AI, democratizing access to state-of-the-art performance. We envision \vlname\ as a foundational resource that empowers the community to build specialized applications and develop more powerful LMMs across diverse tasks through continued scaling.

\bibliographystyle{plainnat}
\bibliography{llava}

@inproceedings{rice,
  title={Region-based Cluster Discrimination for Visual Representation Learning},
  author={Xie, Yin and Yang, Kaicheng and An, Xiang and Wu, Kun and Zhao, Yongle and Deng, Weimo and Ran, Zimin and Wang, Yumeng and Feng, Ziyong and Miles, Roy and others},
  booktitle={ICCV},
  year={2025}
}

@inproceedings{siglip,
  title={Sigmoid loss for language image pre-training},
  author={Zhai, Xiaohua and Mustafa, Basil and Kolesnikov, Alexander and Beyer, Lucas},
  booktitle={ICCV},
  year={2023}
}

@inproceedings{dfn,
  title={Data Filtering Networks},
  author={Fang, Alex and Jose, Albin Madappally and Jain, Amit and Schmidt, Ludwig and Toshev, Alexander and Shankar, Vaishaal},
  booktitle={ICLR},
  year={2023}
}

@article{megatron-lm,
  title={Megatron-LM: Training Multi-Billion Parameter Language Models Using Model Parallelism},
  author={Shoeybi, Mohammad and Patwary, Mostofa and Puri, Raul and LeGresley, Patrick and Casper, Jared and Catanzaro, Bryan},
  journal={arXiv:1909.08053},
  year={2019}
}

@inproceedings{llava1.5,
  title={Improved baselines with visual instruction tuning},
  author={Liu, Haotian and Li, Chunyuan and Li, Yuheng and Lee, Yong Jae},
  booktitle={CVPR},
  year={2024}
}

@inproceedings{llava,
  title={Visual instruction tuning},
  author={Liu, Haotian and Li, Chunyuan and Wu, Qingyang and Lee, Yong Jae},
  booktitle={NeurIPS},
  year={2023}
}

@article{chen2024expanding,
  title={Expanding performance boundaries of open-source multimodal models with model, data, and test-time scaling},
  author={Chen, Zhe and Wang, Weiyun and Cao, Yue and Liu, Yangzhou and Gao, Zhangwei and Cui, Erfei and Zhu, Jinguo and Ye, Shenglong and Tian, Hao and Liu, Zhaoyang and others},
  journal={arXiv:2412.05271},
  year={2024}
}

@article{qwen2.5vl,
  title={Qwen2.5-vl technical report},
  author={Bai, Shuai and Chen, Keqin and Liu, Xuejing and Wang, Jialin and Ge, Wenbin and Song, Sibo and Dang, Kai and Wang, Peng and Wang, Shijie and Tang, Jun and others},
  journal={arXiv:2502.13923},
  year={2025}
}

@misc{llavanext,
    title={LLaVA-NeXT: Improved reasoning, OCR, and world knowledge},
    url={https://llava-vl.github.io/blog/2024-01-30-llava-next/},
    author={Liu, Haotian and Li, Chunyuan and Li, Yuheng and Li, Bo and Zhang, Yuanhan and Shen, Sheng and Lee, Yong Jae},
    year={2024}
}

@article{internvl3,
  title={Internvl3: Exploring advanced training and test-time recipes for open-source multimodal models},
  author={Zhu, Jinguo and Wang, Weiyun and Chen, Zhe and Liu, Zhaoyang and Ye, Shenglong and Gu, Lixin and Tian, Hao and Duan, Yuchen and Su, Weijie and Shao, Jie and others},
  journal={arXiv:2504.10479},
  year={2025}
}

@inproceedings{mmstar,
  title={Are we on the right way for evaluating large vision-language models?},
  author={Chen, Lin and Li, Jinsong and Dong, Xiaoyi and Zhang, Pan and Zang, Yuhang and Chen, Zehui and Duan, Haodong and Wang, Jiaqi and Qiao, Yu and Lin, Dahua and others},
  booktitle={NeurIPS},
  year={2024}
}

@article{mme,
  title={MME: A Comprehensive Evaluation Benchmark for Multimodal Large Language Models},
  author={Fu, Chaoyou and Chen, Peixian and Shen, Yunhang and Qin, Yulei and Zhang, Mengdan and Lin, Xu and Yang, Jinrui and Zheng, Xiawu and Li, Ke and Sun, Xing and others},
  journal={arXiv:2306.13394},
  year={2023}
}

@inproceedings{mmerealworld,
  title={{MME-Realworld: Could your multimodal llm challenge high-resolution real-world scenarios that are difficult for humans?}},
  author={Zhang, Yi-Fan and Zhang, Huanyu and Tian, Haochen and Fu, Chaoyou and Zhang, Shuangqing and Wu, Junfei and Li, Feng and Wang, Kun and Wen, Qingsong and Zhang, Zhang and others},
  booktitle={ICLR},
  year={2025}
}

@inproceedings{seedbench,
  title={Seed-bench: Benchmarking multimodal llms with generative comprehension},
  author={Li, Bohao and Wang, Rui and Wang, Guangzhi and Ge, Yuying and Ge, Yixiao and Shan, Ying},
  booktitle={CVPR},
  year={2024}
}

@inproceedings{infovqa,
  title={Infographicvqa},
  author={Mathew, Minesh and Bagal, Viraj and Tito, Rub{\`e}n and Karatzas, Dimosthenis and Valveny, Ernest and Jawahar, CV},
  booktitle={WACV},
  year={2022}
}

@misc{realworldqa,
  title = {Grok-1.5 Vision Preview: Connecting the digital and physical worlds with our first multimodal model.},
  author = {X.AI Corp.},
  year = {2024},
  howpublished = {\url{https://x.ai/blog/grok-1.5v}}
}

@inproceedings{mathvision,
  title={Measuring multimodal mathematical reasoning with math-vision dataset},
  author={Wang, Ke and Pan, Junting and Shi, Weikang and Lu, Zimu and Zhan, Mingjie and Li, Hongsheng},
  booktitle={NeurIPS},
  year={2024}
}

@inproceedings{mmmupro,
  title={Mmmu-pro: A more robust multi-discipline multimodal understanding benchmark},
  author={Yue, Xiang and Zheng, Tianyu and Ni, Yuansheng and Wang, Yubo and Zhang, Kai and Tong, Shengbang and Sun, Yuxuan and Yin, Ming and Yu, Botao and Zhang, Ge and others},
  booktitle={ACL},
  year={2025}
}

@inproceedings{chartqa,
  title={ChartQA: A Benchmark for Question Answering about Charts with Visual and Logical Reasoning},
  author={Masry, Ahmed and Do, Xuan Long and Tan, Jia Qing and Joty, Shafiq and Hoque, Enamul},
  booktitle={ACL},
  year={2022}
}

@inproceedings{docvqa,
  title={Docvqa: A dataset for vqa on document images},
  author={Mathew, Minesh and Karatzas, Dimosthenis and Jawahar, CV},
  booktitle={WACV},
  year={2021}
}

@inproceedings{charxiv,
  title={Charxiv: Charting gaps in realistic chart understanding in multimodal llms},
  author={Wang, Zirui and Xia, Mengzhou and He, Luxi and Chen, Howard and Liu, Yitao and Zhu, Richard and Liang, Kaiqu and Wu, Xindi and Liu, Haotian and Malladi, Sadhika and others},
  booktitle={NeurIPS},
  year={2024}
}

@inproceedings{mmmu,
  title={Mmmu: A massive multi-discipline multimodal understanding and reasoning benchmark for expert agi},
  author={Yue, Xiang and Ni, Yuansheng and Zhang, Kai and Zheng, Tianyu and Liu, Ruoqi and Zhang, Ge and Stevens, Samuel and Jiang, Dongfu and Ren, Weiming and Sun, Yuxuan and others},
  booktitle={CVPR},
  year={2024}
}

@inproceedings{wemath,
  title={{We-math: Does your large multimodal model achieve human-like mathematical reasoning?}},
  author={Qiao, Runqi and Tan, Qiuna and Dong, Guanting and Wu, Minhui and Sun, Chong and Song, Xiaoshuai and GongQue, Zhuoma and Lei, Shanglin and Wei, Zhe and Zhang, Miaoxuan and others},
  booktitle={ACL},
  year={2025}
}

@inproceedings{mathvista,
  title={Mathvista: Evaluating mathematical reasoning of foundation models in visual contexts},
  author={Lu, Pan and Bansal, Hritik and Xia, Tony and Liu, Jiacheng and Li, Chunyuan and Hajishirzi, Hannaneh and Cheng, Hao and Chang, Kai-Wei and Galley, Michel and Gao, Jianfeng},
  booktitle={ICLR},
  year={2024}
}

@inproceedings{seedbench2,
  title={Seed-bench-2-plus: Benchmarking multimodal large language models with text-rich visual comprehension},
  author={Li, Bohao and Ge, Yuying and Chen, Yi and Ge, Yixiao and Zhang, Ruimao and Shan, Ying},
  booktitle={ICLR},
  year={2024}
}

@inproceedings{wang2025open,
  title={Open-Qwen2VL: Compute-Efficient Pre-Training of Fully-Open Multimodal LLMs on Academic Resources},
  author={Wang, Weizhi and Tian, Yu and Yang, Linjie and Wang, Heng and Yan, Xifeng},
  booktitle={COLM},
  year={2025}
}

@inproceedings{molmo,
  title={Molmo and pixmo: Open weights and open data for state-of-the-art vision-language models},
  author={Deitke, Matt and Clark, Christopher and Lee, Sangho and Tripathi, Rohun and Yang, Yue and Park, Jae Sung and Salehi, Mohammadreza and Muennighoff, Niklas and Lo, Kyle and Soldaini, Luca and others},
  booktitle={CVPR},
  year={2025}
}

@article{llavaov,
  title={Llava-onevision: Easy visual task transfer},
  author={Li, Bo and Zhang, Yuanhan and Guo, Dong and Zhang, Renrui and Li, Feng and Zhang, Hao and Zhang, Kaichen and Zhang, Peiyuan and Li, Yanwei and Liu, Ziwei and others},
  journal={TMLR},
  year={2025}
}

@misc{coyo700m,
  title= {COYO-700M: Image-Text Pair Dataset},
  author= {Byeon, Minwoo and Park, Beomhee and Kim, Haecheon and Lee, Sungjun and Baek, Woonhyuk and Kim, Saehoon},
  year = {2022},
  howpublished= {\url{https://github.com/kakaobrain/coyo-dataset}},
}

@inproceedings{obelics,
  title={Obelics: An open web-scale filtered dataset of interleaved image-text documents},
  author={Lauren{\c{c}}on, Hugo and Saulnier, Lucile and Tronchon, L{\'e}o and Bekman, Stas and Singh, Amanpreet and Lozhkov, Anton and Wang, Thomas and Karamcheti, Siddharth and Rush, Alexander and Kiela, Douwe and others},
  booktitle={NeurIPS},
  year={2023}
}

@inproceedings{datacomp,
  title={Datacomp: In search of the next generation of multimodal datasets},
  author={Gadre, Samir Yitzhak and Ilharco, Gabriel and Fang, Alex and Hayase, Jonathan and Smyrnis, Georgios and Nguyen, Thao and Marten, Ryan and Wortsman, Mitchell and Ghosh, Dhruba and Zhang, Jieyu and others},
  booktitle={NeurIPS},
  year={2023}
}

@article{laioncn,
  author= {Jiaxing Zhang and Ruyi Gan and Junjie Wang and Yuxiang Zhang and Lin Zhang and Ping Yang and Xinyu Gao and Ziwei Wu and Xiaoqun Dong and Junqing He and Jianheng Zhuo and Qi Yang and Yongfeng Huang and Xiayu Li and Yanghan Wu and Junyu Lu and Xinyu Zhu and Weifeng Chen and Ting Han and Kunhao Pan and Rui Wang and Hao Wang and Xiaojun Wu and Zhongshen Zeng and Chongpei Chen},
  title = {Fengshenbang 1.0: Being the Foundation of Chinese Cognitive Intelligence},
  journal={arXiv:2209.02970},
  year= {2022}
}

@article{imagenet,
  title={Imagenet large scale visual recognition challenge},
  author={Russakovsky, Olga and Deng, Jia and Su, Hao and Krause, Jonathan and Satheesh, Sanjeev and Ma, Sean and Huang, Zhiheng and Karpathy, Andrej and Khosla, Aditya and Bernstein, Michael and others},
  journal={IJCV},
  year={2015},
}

@inproceedings{sam,
  title={Segment anything},
  author={Kirillov, Alexander and Mintun, Eric and Ravi, Nikhila and Mao, Hanzi and Rolland, Chloe and Gustafson, Laura and Xiao, Tete and Whitehead, Spencer and Berg, Alexander C and Lo, Wan-Yen and others},
  booktitle={ICCV},
  year={2023}
}

@inproceedings{mint,
  title={Mint: Evaluating llms in multi-turn interaction with tools and language feedback},
  author={Wang, Xingyao and Wang, Zihan and Liu, Jiateng and Chen, Yangyi and Yuan, Lifan and Peng, Hao and Ji, Heng},
  booktitle={ICLR},
  year={2024}
}

@inproceedings{zero,
  title={CCMB: A Large-scale Chinese Cross-modal Benchmark},
  author={Xie, Chunyu and Cai, Heng and Li, Jincheng and Kong, Fanjing and Wu, Xiaoyu and Song, Jianfei and Morimitsu, Henrique and Yao, Lin and Wang, Dexin and Zhang, Xiangzheng and others},
  booktitle={ACMMM},
  year={2023}
}

@article{wang2024qwen2,
  title={Qwen2-vl: Enhancing vision-language model's perception of the world at any resolution},
  author={Wang, Peng and Bai, Shuai and Tan, Sinan and Wang, Shijie and Fan, Zhihao and Bai, Jinze and Chen, Keqin and Liu, Xuejing and Wang, Jialin and Ge, Wenbin and others},
  journal={arXiv:2409.12191},
  year={2024}
}

@inproceedings{metaclip,
  title={Demystifying clip data},
  author={Xu, Hu and Xie, Saining and Tan, Xiaoqing Ellen and Huang, Po-Yao and Howes, Russell and Sharma, Vasu and Li, Shang-Wen and Ghosh, Gargi and Zettlemoyer, Luke and Feichtenhofer, Christoph},
  booktitle={ICLR},
  year={2024}
}

@inproceedings{clip,
      title={Learning Transferable Visual Models From Natural Language Supervision}, 
      author={Alec Radford and Jong Wook Kim and Chris Hallacy and Aditya Ramesh and Gabriel Goh and Sandhini Agarwal and Girish Sastry and Amanda Askell and Pamela Mishkin and Jack Clark and Gretchen Krueger and Ilya Sutskever},
      booktitle={ICML},
      year={2021},
}

@inproceedings{siglipv2,
      title={SigLIP 2: Multilingual Vision-Language Encoders with Improved Semantic Understanding, Localization, and Dense Features}, 
      author={Michael Tschannen and Alexey Gritsenko and Xiao Wang and Muhammad Ferjad Naeem and Ibrahim Alabdulmohsin and Nikhil Parthasarathy and Talfan Evans and Lucas Beyer and Ye Xia and Basil Mustafa and Olivier Hénaff and Jeremiah Harmsen and Andreas Steiner and Xiaohua Zhai},
      year={2025},
      booktitle={arXiv:2502.14786},
}

@inproceedings{fini2025multimodal,
    title={Multimodal Autoregressive Pre-training of Large Vision Encoders},
    author={Enrico Fini and Mustafa Shukor and Xiujun Li and Philipp Dufter and Michal Klein and David Haldimann and Sai Aitharaju and Victor Guilherme Turrisi da Costa and Louis Béthune and Zhe Gan and Alexander T Toshev and Marcin Eichner and Moin Nabi and Yinfei Yang and Joshua M. Susskind and Alaaeldin El-Nouby},
    booktitle={CVPR},
    year={2025},
}

@article{ocrbench,
   title={OCRBench: on the hidden mystery of OCR in large multimodal models},
   journal={Science China Information Sciences},
   author={Liu, Yuliang and Li, Zhang and Huang, Mingxin and Yang, Biao and Yu, Wenwen and Li, Chunyuan and Yin, Xu-Cheng and Liu, Cheng-Lin and Jin, Lianwen and Bai, Xiang},
   year={2024},
}

@inproceedings{ai2d,
  title={A diagram is worth a dozen images},
  author={Kembhavi, Aniruddha and Salvato, Mike and Kolve, Eric and Seo, Minjoon and Hajishirzi, Hannaneh and Farhadi, Ali},
  booktitle={ECCV},
  year={2016}
}

@inproceedings{paiss2023teaching,
  title={Teaching clip to count to ten},
  author={Paiss, Roni and Ephrat, Ariel and Tov, Omer and Zada, Shiran and Mosseri, Inbar and Irani, Michal and Dekel, Tali},
  booktitle={ICCV},
  year={2023}
}

@inproceedings{li2025vl,
  title={VL-RewardBench: A Challenging Benchmark for Vision-Language Generative Reward Models},
  author={Li, Lei and Wei, Yuancheng and Xie, Zhihui and Yang, Xuqing and Song, Yifan and Wang, Peiyi and An, Chenxin and Liu, Tianyu and Li, Sujian and Lin, Bill Yuchen and others},
  booktitle={CVPR},
  year={2025}
}

@inproceedings{wu2024v,
  title={V*: Guided visual search as a core mechanism in multimodal llms},
  author={Wu, Penghao and Xie, Saining},
  booktitle={CVPR},
  year={2024}
}

@inproceedings{tong2024cambrian,
  title={Cambrian-1: A fully open, vision-centric exploration of multimodal llms},
  author={Tong, Peter and Brown, Ellis and Wu, Penghao and Woo, Sanghyun and IYER, Adithya Jairam Vedagiri and Akula, Sai Charitha and Yang, Shusheng and Yang, Jihan and Middepogu, Manoj and Wang, Ziteng and others},
  booktitle={NeurIPS},
  year={2024}
}

@inproceedings{mlcd,
  title={Multi-label cluster discrimination for visual representation learning},
  author={An, Xiang and Yang, Kaicheng and Dai, Xiangzi and Feng, Ziyong and Deng, Jiankang},
  booktitle={ECCV},
  year={2024},
}

@article{gemini2.5,
  title={Gemini 2.5: Pushing the frontier with advanced reasoning, multimodality, long context, and next generation agentic capabilities},
  author={Comanici, Gheorghe and Bieber, Eric and Schaekermann, Mike and Pasupat, Ice and Sachdeva, Noveen and Dhillon, Inderjit and Blistein, Marcel and Ram, Ori and Zhang, Dan and Rosen, Evan and others},
  journal={arXiv:2507.06261},
  year={2025}
}

@article{seedv11.5,
  title={Seed1. 5-vl technical report},
  author={Guo, Dong and Wu, Faming and Zhu, Feida and Leng, Fuxing and Shi, Guang and Chen, Haobin and Fan, Haoqi and Wang, Jian and Jiang, Jianyu and Wang, Jiawei and others},
  journal={arXiv:2505.07062},
  year={2025}
}

@misc{qwen3,
      title={Qwen3 Technical Report}, 
      author={Qwen Team},
      year={2025},
      journal={arXiv:2505.09388},
}

@inproceedings{finevision2025,
      title={FineVision: Open Data Is All You Need}, 
      author={Luis Wiedmann and Orr Zohar and Amir Mahla and Xiaohan Wang and Rui Li and Thibaud Frere and Leandro von Werra and Aritra Roy Gosthipaty and Andrés Marafioti},
      year={2025},
      booktitle={arXiv:2510.17269},
}

@inproceedings{lmms_eval,
      title={LMMs-Eval: Reality Check on the Evaluation of Large Multimodal Models}, 
      author={Kaichen Zhang and Bo Li and Peiyuan Zhang and Fanyi Pu and Joshua Adrian Cahyono and Kairui Hu and Shuai Liu and Yuanhan Zhang and Jingkang Yang and Chunyuan Li and Ziwei Liu},
      booktitle={ACL},
      year={2025},
}

@inproceedings{
yuan2024virl39k,
title={VL-Rethinker: Incentivizing Self-Reflection of Vision-Language Models with Reinforcement Learning},
author={Haozhe Wang and Chao Qu and Zuming Huang and Wei Chu and Fangzhen Lin and Wenhu Chen},
booktitle={NeurIPS},
year={2025},
}

@inproceedings{
zhou2024webcode2m,
title={Web2Code: A Large-scale Webpage-to-Code Dataset and Evaluation Framework for Multimodal {LLM}s},
author={Sukmin Yun and Haokun Lin and Rusiru Thushara and Mohammad Qazim Bhat and Yongxin Wang and Zutao Jiang and Mingkai Deng and Jinhong Wang and Tianhua Tao and Junbo Li and Haonan Li and Preslav Nakov and Timothy Baldwin and Zhengzhong Liu and Eric P. Xing and Xiaodan Liang and Zhiqiang Shen},
booktitle={NeurIPS},
year={2024},
}

@article{chen2024revisiting,
  title={Revisiting Referring Expression Comprehension Evaluation in the Era of Large Multimodal Models},
  author={Chen, Jierun and Wei, Fangyun and Zhao, Jinjing and Song, Sizhe and Wu, Bohuai and Peng, Zhuoxuan and Chan, S.-H. Gary},
  journal={arXiv:2406.16866},
  year={2024}
}

@inproceedings{
sarch2025vigorl,
title={Grounded Reinforcement Learning for Visual Reasoning},
author={Gabriel Herbert Sarch and Snigdha Saha and Naitik Khandelwal and Ayush Jain and Michael J. Tarr and Aviral Kumar and Katerina Fragkiadaki},
booktitle={NeurIPS},
year={2025},
}

@inproceedings{li2025unisvg,
author = {Li, Jinke and Yu, Jiarui and Wei, Chenxing and Dong, Hande and Lin, Qiang and Yang, Liangjing and Wang, Zhicai and Hao, Yanbin},
title = {UniSVG: A Unified Dataset for Vector Graphic Understanding and Generation with Multimodal Large Language Models},
booktitle = {ACMMM},
year={2025}
}

@article{shao2024deepseekmath,
  title={Deepseekmath: Pushing the limits of mathematical reasoning in open language models},
  author={Shao, Zhihong and Wang, Peiyi and Zhu, Qihao and Xu, Runxin and Song, Junxiao and Bi, Xiao and Zhang, Haowei and Zhang, Mingchuan and Li, YK and Wu, Yang and others},
  journal={arXiv:2402.03300},
  year={2024}
}

@inproceedings{kembhavi2016diagram,
  title={A Diagram Is Worth A Dozen Images},
  author={Kembhavi, Aniruddha and Salvato, Mike and Kolve, Eric and Seo, Minjoon and Hajishirzi, Hannaneh and Farhadi, Ali},
  booktitle={ECCV},
  year={2016}
}

@inproceedings{mathew2022infographicvqa,
  title={InfographicVQA},
  author={Mathew, Minesh and Bagal, Viraj and Karatzas, Dimosthenis and Jawahar, C. V.},
  booktitle={WACV},
  year={2022}
}

@inproceedings{
fu2025areal,
title={{AREAL}: A Large-Scale Asynchronous Reinforcement Learning System for Language Reasoning},
author={Wei Fu and Jiaxuan Gao and Xujie Shen and Chen Zhu and Zhiyu Mei and Chuyi He and Shusheng Xu and Guo Wei and Jun Mei and WANG JIASHU and Tongkai Yang and Binhang Yuan and Yi Wu},
booktitle={NeurIPS},
year={2025},
}

\newpage
\clearpage
\appendix

\section{LLaVA-OV-1.5 vs. Qwen2.5-VL with Same LLM}
\begin{figure}[h!]
    \centering
    \includegraphics[width=0.85\textwidth]{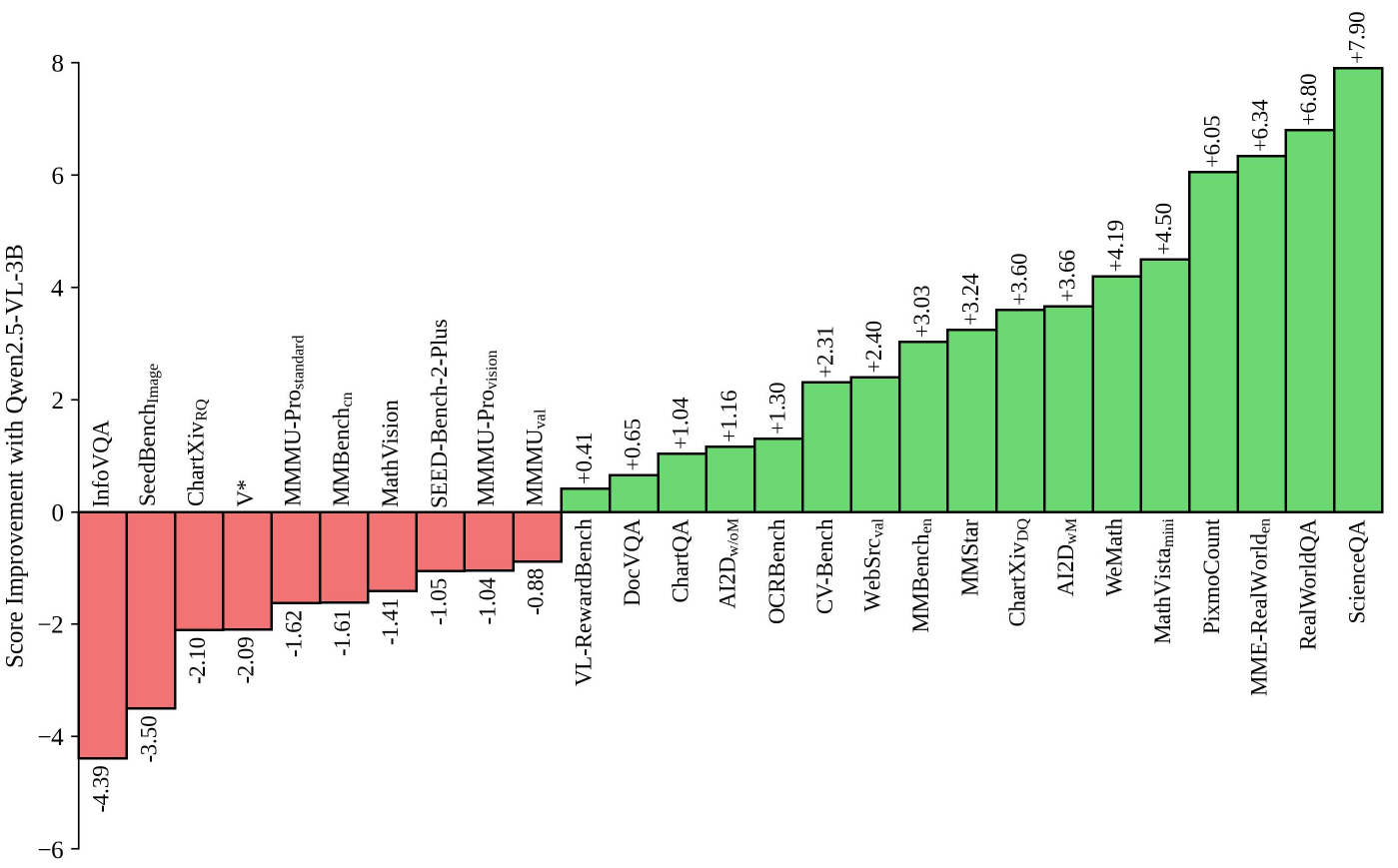}
    \vspace{-2mm}
    \caption{Comparison between LLaVA-OV-1.5-3B  and Qwen2.5-VL-3B model on public datasets using the same LLM for both evaluations.}
    \vspace{-3mm}
    \label{fig:3b}
\end{figure}

To enable a fair comparison with Qwen2.5-VL, we train LLaVA-Onevision-1.5-3B based on Qwen2.5-3B-Instruct. As shown in Fig.~\ref{fig:3b}, LLaVA-Onevision-1.5-3B also demonstrates superior performance, achieving better results on 17 out of 27 downstream benchmarks. 

\begin{figure}[h!]
    \centering
    \includegraphics[width=\textwidth]{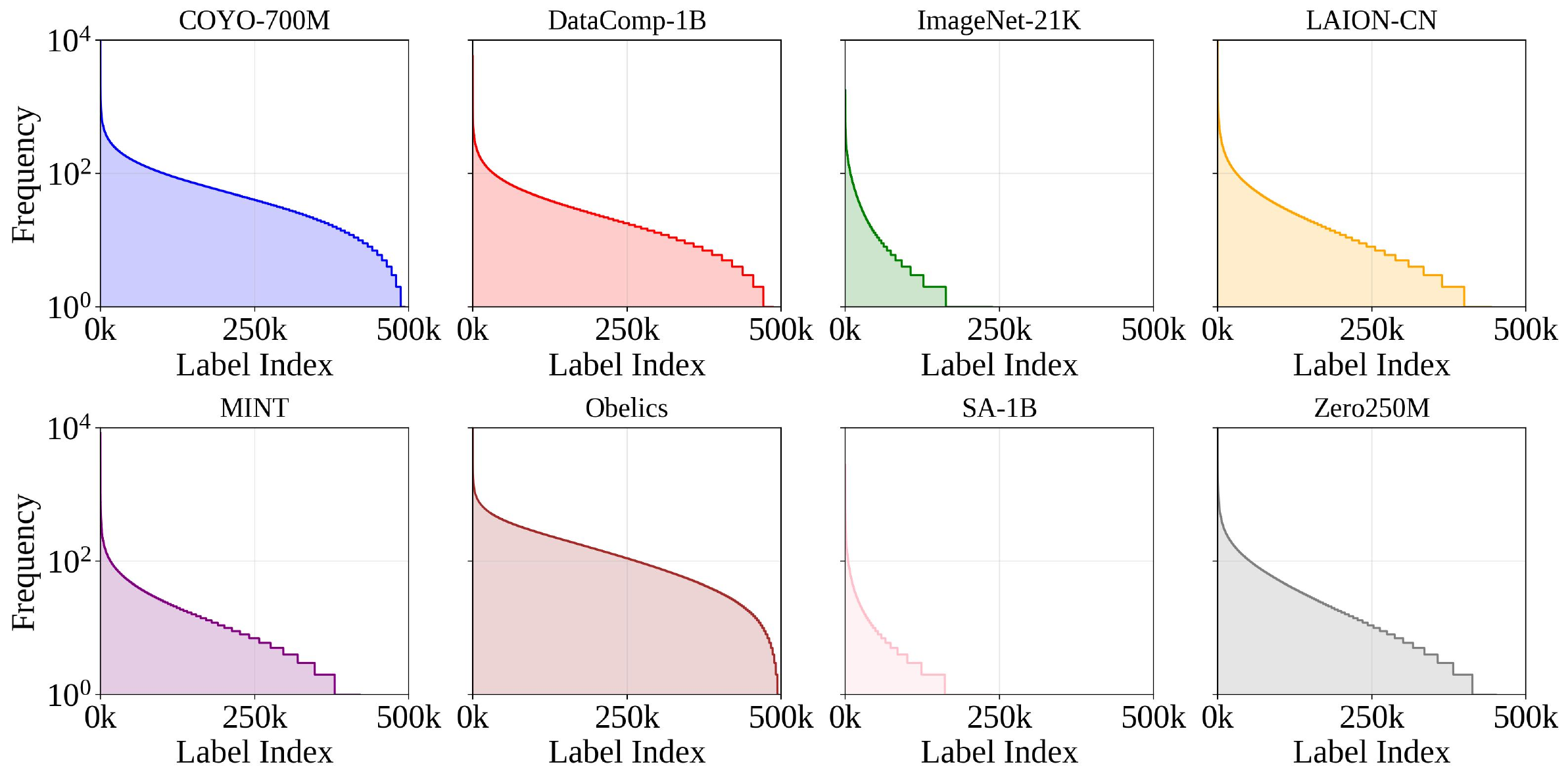}
    \vspace{-5mm}
    \caption{Original concept distributions across eight common vision datasets used in the LLaVA-OneVision-1.5-Mid-Traning dataset.}
    \label{fig:concept_dis}
    \vspace{-3mm}
\end{figure}

\section{Mid-Training Data: Concept Distribution and Top 50 Topics}

Fig.~\ref{fig:concept_dis} presents the raw concept distributions across eight common vision datasets used in the LLaVA-OneVision-1.5-Mid-Traning dataset. All sources exhibit a pronounced long‑tail bias, indicating that the original data are far from comprehensive. Obelics displays the broadest and most uniform distribution (slowest tail decay), while others, such as ImageNet-21K and SA-1B, cover fewer concepts with higher frequency concentration. To characterize the semantic space of the mid-training mixture, we apply topic modeling to associated texts and extract the 50 most salient topics (Tab. \ref{tab:topics}). These span diverse domains, including wildlife, apparel, cuisine, interiors, engineering, electronics, healthcare, WebUI, and cultural activities, offering an interpretable summary of data coverage. This analysis confirms Obelics as the most comprehensive source and highlights coverage gaps in other datasets, informing subsequent data curation.

{\begin{longtable} {p{5cm} p{10cm}}
\caption{Topic Modeling Results (50 Topics)}
\label{tab:topics} \\ 
\toprule 
\textbf{Concept} & \textbf{Related words} \\
\midrule
\endfirsthead

\multicolumn{2}{c}%
{{\tablename\ \thetable{} -- continued from previous page}} \\
\midrule
\textbf{Concept} & \textbf{Related words} \\
\midrule
\endhead

\multicolumn{2}{r}{{Continued on next page}} \\
\endfoot

\bottomrule
\endlastfoot

Wildlife & feathers, wildlife, beak, branch, birds, habitat, scales, plumage, claws, feather, behavior, spots, watchers, snout, species, creature, paws, limbs, flight, drawing \\
\midrule
Jewelry and Aviation & jewelry, beads, airplane, aircraft, aviation, plane, sparkle, bracelet, flight, stones, elegance, pendant, accessory, loop, diamonds, gemstones, gemstone, pearl, landing, shine \\
\midrule
Internet Humor & humor, phrase, eco, baby, parents, sustainability, frustration, references, meme, sentiment, surprise, mother, farm, disney, twist, internet, cartoonish, novelty, reaction, drawing \\
\midrule
Education and Learning & slide, educators, learners, bullet, education, learning, classroom, study, math, guide, student, resource, writing, textbook, skills, problem, publisher, question, list, questions \\
\midrule
Watches and Accessories & case, strap, watch, wristwatch, hour, brim, crown, markers, dial, barrel, numerals, grip, accessory, clock, luxury, gun, minute, buckle, rifle, stitching \\
\midrule
Gaming and Sports & games, gamers, baseball, armor, jerseys, fi, franchise, sword, basketball, horror, competition, sport, court, gaming, anime, athleticism, readiness, arena, athletes, flames \\
\midrule
Certification and Professionalism & stamp, clients, certification, seal, colleagues, stamps, standards, calligraphy, headshot, friendliness, compliance, certifications, profiles, mark, envelope, creases, impression, identification, trustworthiness, german \\
\midrule
Vehicles and Racing & seat, wheel, tires, grille, truck, license, headlights, speed, driver, track, rims, quarter, mirrors, race, seats, racing, train, transportation, three-quarter, windshield \\
\midrule
Architectural and Mechanical Plans & grooves, architects, drawings, plan, drawing, tire, hub, roller, wheel, tread, plans, coaster, distances, blueprint, 2023, rubber, planners, amusement, skateboard, traction \\
\midrule
Ceremonies and Achievement & achievement, pride, ceremony, award, trophy, university, campus, awards, skeleton, victory, graduation, bones, success, achievements, alumni, casino, medal, certificate, dinosaur, tuxedo \\
\midrule
Portraiture and Makeup & finger, eyebrows, jawline, cheek, makeup, cheeks, portraiture, index, nails, individual, strands, thumb, beer, cheekbones, brow, bangs, ear, shot, stubble, creases \\
\midrule
Software and Development & software, computer, developers, options, system, user, code, management, file, version, application, menu, fields, input, process, arrows, programming, files, 3d, interfaces \\
\midrule
Pets and Footwear & dog, footwear, owners, cat, shoe, laces, toe, sole, heel, paws, tongue, sneakers, soles, mesh, ankle, rubber, foot, dogs, toes, straps \\
\midrule
Typography and Design & sans, gradients, letter, variations, symbolism, forms, spacing, thickness, variation, meaning, bold, alignment, curves, designers, representation, emotions, rectangle, curve, knowledge, trademark \\
\midrule
Cuisine and Cooking & cooks, meal, ingredients, dish, freshness, vegetables, cooking, sauce, fruit, slices, pieces, herbs, spoon, meat, cuisine, crispy, slice, rice, fruits, chocolate \\
\midrule
Home Interiors & coffee, lamp, shelf, living, vase, bedroom, cabinet, pillows, comfort, rug, countertop, cup, homeowners, sofa, pillow, cabinets, counter, flooring, couch, shade \\
\midrule
Skincare and Packaging & skincare, liquid, lid, wine, container, ingredients, tube, supplements, screw, premium, luxury, benefits, transparency, labeling, jar, solutions, supplement, powder, oil, spray \\
\midrule
Childhood and Holidays & parents, toy, holiday, baby, gift, toys, polka, kids, christmas, decorations, ribbon, bear, caregivers, fun, charm, decoration, greeting, rabbit, birthday, dot \\
\midrule
Social Interactions & pen, smiles, postures, mid-20th-century, discussion, celebrity, hairstyles, collage, interactions, collaboration, jackets, notebook, -century, fourth, gestures, cinema, teamwork, interview, ties, plaid \\
\midrule
Literature and Romance & love, spine, intimacy, works, couple, publisher, affection, novel, romance, proximity, moments, bond, hardcover, 19th-century, pages, edition, poetry, drama, collection, covers \\
\midrule
Inspirational Quotes & quote, self, positivity, inspiration, resilience, philosophy, quotes, hope, help, introspection, journey, phrase, attribution, things, weight, motivation, freedom, contemplation, love, vulnerability \\
\midrule
Waterside Leisure & beach, shore, ripples, relaxation, sand, roofs, boat, sea, pool, bridge, shoreline, turquoise, landscapes, boats, river, lake, deck, skyline, trunks, leisure \\
\midrule
Engineering and Mechanics & component, engineers, holes, mechanics, diy, engineering, technicians, screws, hardware, hole, specifications, wires, electronics, manufacturers, steel, manufacturing, manufacturer, circuit, grip, ends \\
\midrule
Retail and Pricing & price, package, shoppers, 10, barcode, promotion, pack, store, 20, sale, 100, tag, info, shipping, 50, medication, 30, sales, discount, quantity \\
\midrule
Technology and Innovation & innovation, network, lightning, connectivity, bolt, globe, connections, nodes, ideas, cloud, gears, communication, networks, bulb, intelligence, integration, lightbulb, concept, telecommunications, networking \\
\midrule
Industrial and Manufacturing & workers, machine, machinery, manufacturing, storage, safety, fish, facility, factory, workshop, pipes, warehouse, maintenance, logistics, task, site, fins, warning, steel, production \\
\midrule
Trust and Publication & trust, shield, reliability, studies, china, stakeholders, clients, partners, publication, translation, institution, publisher, public, tradition, scholars, formality, stability, organizations, strength, academics \\
\midrule
Healthcare and Fitness & healthcare, fitness, patients, training, wellness, exercise, hospital, practice, gym, strength, clinic, muscles, shirtless, support, lab, treatment, stethoscope, muscle, weight, doctor \\
\midrule
Museum Artifacts & motifs, sculpture, bronze, museum, statue, coin, folds, engravings, artifacts, skull, scroll, item, carvings, coins, gallery, pedestal, scratches, artifact, elegance, artwork \\
\midrule
Dining and Hospitality & tables, restaurant, café, experience, spot, counter, customers, seating, locals, diners, drinks, shop, establishment, patrons, menu, casual, fixtures, hats, market, drink \\
\midrule
Real Estate & estate, tiles, homeowners, shrubs, bathroom, lawn, railings, bushes, porch, property, driveway, staircase, bricks, houses, panes, yard, homebuyers, neighborhood, chimney, concrete \\
\midrule
Music and Geography & guitar, musicians, concert, instrument, instruments, country, artist, strings, regions, musician, roads, singer, drum, song, locations, geography, land, performers, performances, maps \\
\midrule
Entomology and Containers & lid, antennae, container, insect, butterfly, spots, shell, spines, wing, ridges, candle, storage, rim, insects, hairs, abdomen, entomology, bee, iridescent, observation \\
\midrule
Apparel Design & apparel, pocket, neckline, pockets, waist, garment, fit, cuffs, zipper, hip, belt, tag, straps, torso, cotton, crew, seams, hem, hood, knee \\
\midrule
Crafting and DIY & crafters, craft, supplies, crafts, diy, artists, projects, creativity, lab, textiles, textile, laboratory, project, pieces, motifs, stationery, pencil, thread, tactile, workspace \\
\midrule
Data and Chemistry & graph, comparison, values, axis, trends, analysis, chart, x-, analysts, market, chemistry, metrics, chemical, visualization, graphs, 2d, years, investors, axes, groups \\
\midrule
Electronic Devices & devices, smartphone, computer, keyboard, laptop, lens, monitor, cable, indicator, case, audio, keys, speaker, tablet, usb, ports, security, workspace, electronics, solutions \\
\midrule
Recreation and Law & police, law, sheet, superhero, notes, golf, enforcement, club, course, officer, officers, cape, notation, copyright, costume, security, piano, personnel, swing, musicians \\
\midrule
Web and UI & navigation, app, user, header, links, webpage, search, smartphone, screenshot, web, options, link, yuan, tabs, battery, url, post, site, menu, email \\
\midrule
Space and Music Media & moon, earth, exploration, record, wonder, vinyl, planet, purples, astronomy, records, sphere, thirds, crescent, two-thirds, landmasses, herbarium, rocket, oceans, cloud, magic \\
\midrule
Business and Finance Documents & bullet, info, finance, header, management, money, address, documents, topics, question, headers, description, list, headings, job, email, fields, slide, dollar, newspaper \\
\midrule
Weddings and Traditions & wedding, lace, embroidery, dance, festival, dresses, indian, traditions, bouquet, occasion, gown, couple, bride, costumes, couples, weddings, outfits, tradition, headscarf, veil \\
\midrule
Transportation and Fiction & horse, ship, bike, mystery, bicycle, soldiers, handlebars, thriller, hull, motorcycle, crime, smoke, genre, rider, genres, cyclists, drama, horses, war, dramas \\
\midrule
Botany and Gardening & stems, veins, cluster, stem, blooms, bloom, gardeners, soil, gardening, tips, clusters, bark, buds, trunk, stamens, botany, canopy, blossoms, fishing, droplets \\
\midrule
Vast Landscapes & terrain, peaks, landscapes, desert, expanse, slopes, range, formations, vastness, isolation, grandeur, peak, hikers, valley, stones, mist, appeals, earth, slope, awe \\
\midrule
Biological Diagrams & diagram, arrows, biology, research, documentation, anatomy, cell, diagrams, purposes, cells, blood, flow, study, emergency, specimen, scientists, tissue, understanding, measurements, response \\
\midrule
Religious Architecture & carvings, grandeur, church, robe, dome, reverence, christian, robes, temple, site, statue, landmark, spires, arches, european, towers, statues, arch, castle, spire \\
\midrule
Time and Health Data & dates, days, calendar, pm, masks, covid-19, times, info, week, month, ray, pandemic, schedule, daily, 1st, sunday, flyer, dna, 2020, helix \\
\midrule
Public Speaking & speech, bubble, press, speaker, podium, politics, pin, government, formality, communication, mark, bubbles, microphones, announcement, lapel, u.s., gesture, exclamation, gravity, gestures \\
\midrule
Urban Streetscape & signage, mid-morning, -morning, sidewalk, pavement, parking, pole, traffic, hotel, poles, store, locals, awning, lot, streetlights, pedestrians, station, residents, storefront, japan \\

\end{longtable}}

\section{Contributors}

\begin{minipage}[t]{0.48\textwidth}
\raggedright
\textbf{Contributors}

\begin{itemize}[noitemsep,topsep=0pt,leftmargin=*]
\item Xiang An
\item Yin Xie
\item Kaicheng Yang
\item Wenkang Zhang
\item Xiuwei Zhao
\item Zheng Cheng
\item Yirui Wang
\item Songcen Xu
\item Changrui Chen
\item Didi Zhu
\item Chunsheng Wu
\item Huajie Tan
\item Chunyuan Li
\item Jing Yang
\item Jie Yu
\item Xiyao Wang
\item Bin Qin
\item Yumeng Wang
\item Zizhen Yan

\end{itemize}
\end{minipage}\hfill
\begin{minipage}[t]{0.48\textwidth}
\raggedright
\textbf{Project Leaders}

\begin{itemize}[noitemsep,topsep=0pt,leftmargin=*]
\item Ziyong Feng
\item Ziwei Liu
\item Bo Li
\item Jiankang Deng
\end{itemize}
\end{minipage}

\end{document}